\documentclass[lettersize,journal]{IEEEtran}

\usepackage{amsmath,amsfonts}
\usepackage{algorithmic}
\usepackage{algorithm}
\usepackage{array}
\usepackage{textcomp}
\usepackage{stfloats}
\usepackage{url}
\usepackage{verbatim}
\usepackage{graphicx}
\usepackage{cite}

\usepackage{adjustbox}
\usepackage{booktabs}
\usepackage{colortbl}
\usepackage{multirow}
\usepackage{subcaption}
\usepackage{xcolor}
\usepackage[normalem]{ulem}
\useunder{\uline}{\ul}{}

\usepackage{amsfonts,amssymb}
\usepackage{tabularray}
\usepackage{makecell}
\usepackage{placeins}
\usepackage{balance}

\newif\ifshowcomments
\showcommentstrue

\ifshowcomments

\newcommand{\TODO}[1]{{\color{red}{[TODO: #1]}}}

\newcommand{\revised}[1]{{\color[rgb]{0.0,0.0,1.0}{#1}}}

\else
\newcommand{\TODO}[1]{}
\newcommand{\revised}[1]{}
\fi

\begin{document}

\title{{Video Instance Shadow Detection \\ Under the Sun and Sky}}

\author{Zhenghao Xing, Tianyu Wang, Xiaowei Hu,~\IEEEmembership{Member,~IEEE}, Haoran Wu, \\ Chi-Wing Fu,~\IEEEmembership{Member,~IEEE}, and Pheng-Ann Heng,~\IEEEmembership{Senior Member,~IEEE}
\thanks{Z. Xing, T. Wang, H. Wu, C.-W. Fu and P.-A. Heng are with the Department of Computer Science and Engineering, The Chinese University of Hong Kong, Hong Kong SAR, China.}
\thanks{X. Hu is with the Shanghai Artificial Intelligence Laboratory, Shanghai, China.}
\thanks{Corresponding author: Xiaowei Hu (e-mail: huxiaowei@pjlab.org.cn).}
}

\markboth{IEEE Transactions on Image Processing}%
{Shell \MakeLowercase{\textit{et al.}}: A Sample Article Using IEEEtran.cls for IEEE Journals}

\maketitle

\begin{abstract}

Instance shadow detection, crucial for applications such as photo editing and light direction estimation, has undergone significant advancements in predicting shadow instances, object instances, and their associations. 
The extension of this task to videos presents challenges in annotating diverse video data and addressing complexities arising from occlusion and temporary disappearances within associations. 
In response to these challenges, we introduce ViShadow, a semi-supervised video instance shadow detection framework that leverages both labeled image data and unlabeled video data for training. 
ViShadow features a two-stage training pipeline: the first stage, utilizing labeled image data, identifies shadow and object instances through contrastive learning for cross-frame pairing. The second stage employs unlabeled videos, incorporating an associated cycle consistency loss to enhance tracking ability. 
A retrieval mechanism is introduced to manage temporary disappearances, ensuring tracking continuity. 
The SOBA-VID dataset, comprising unlabeled training videos and labeled testing videos, along with the SOAP-VID metric, is introduced for the quantitative evaluation of VISD solutions. 
The effectiveness of ViShadow is further demonstrated through various video-level applications such as video inpainting, instance cloning, shadow editing, and text-instructed shadow-object manipulation.


\end{abstract}

\begin{figure*}[tp]
  \centering
  \includegraphics[width=.99\textwidth]{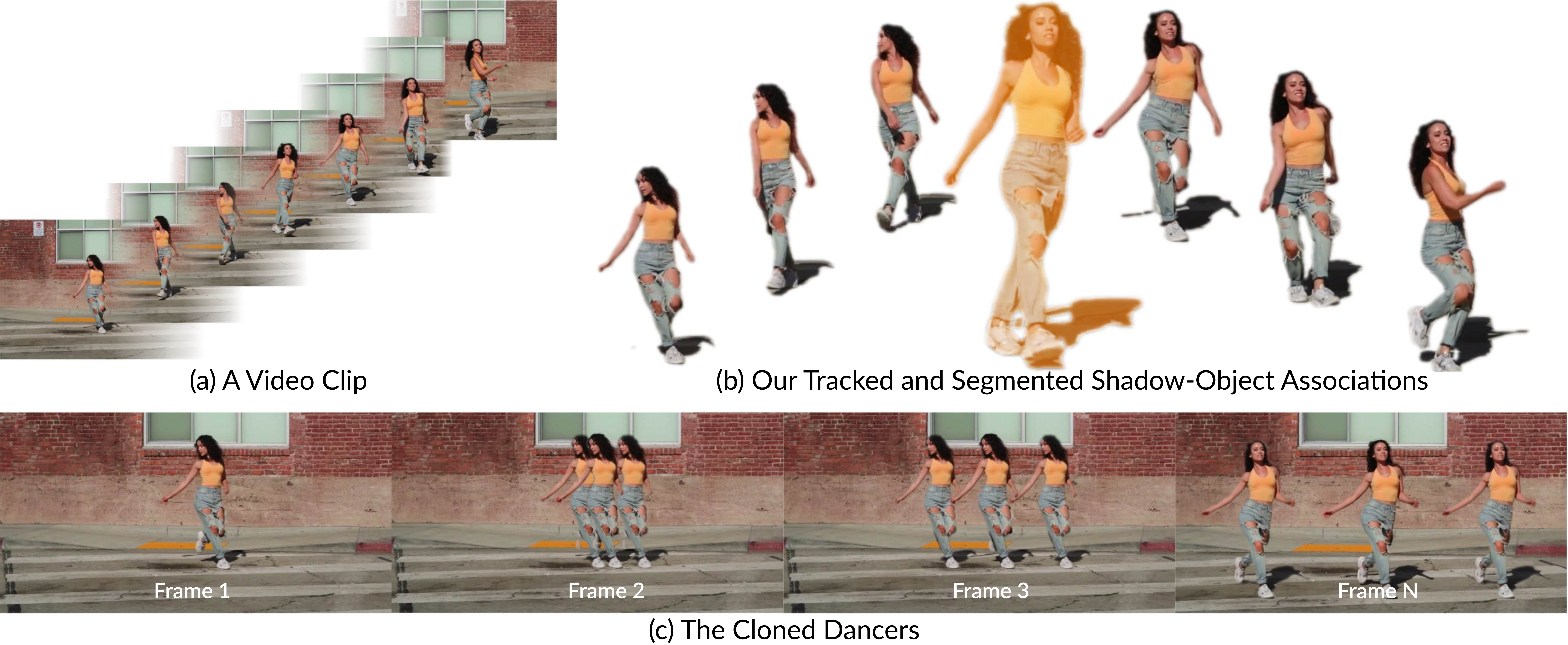}
  \captionof{figure}{Based on the video clip presented in (a), our ViShadow framework demonstrates a robust ability to detect, segment, associate, and track the dancer along with her shadow (b). This capability facilitates a range of applications, such as duplicating the dancer with her shadow to create captivating visual effects (c).
  }
  \label{fig:teaser}
\end{figure*}


\section{Introduction}
\label{sec:intro}

{Instance shadow detection~\cite{wang2020instance} is a crucial task involving the identification of shadows alongside their associated objects in images. 
This task predicts shadow instances, object instances, and the relationships between them by taking the single image as input. }
Its applications span various domains, including photo editing and light direction estimation.

\begin{figure*}[tp]
    \centering
    \includegraphics[width=1\linewidth]{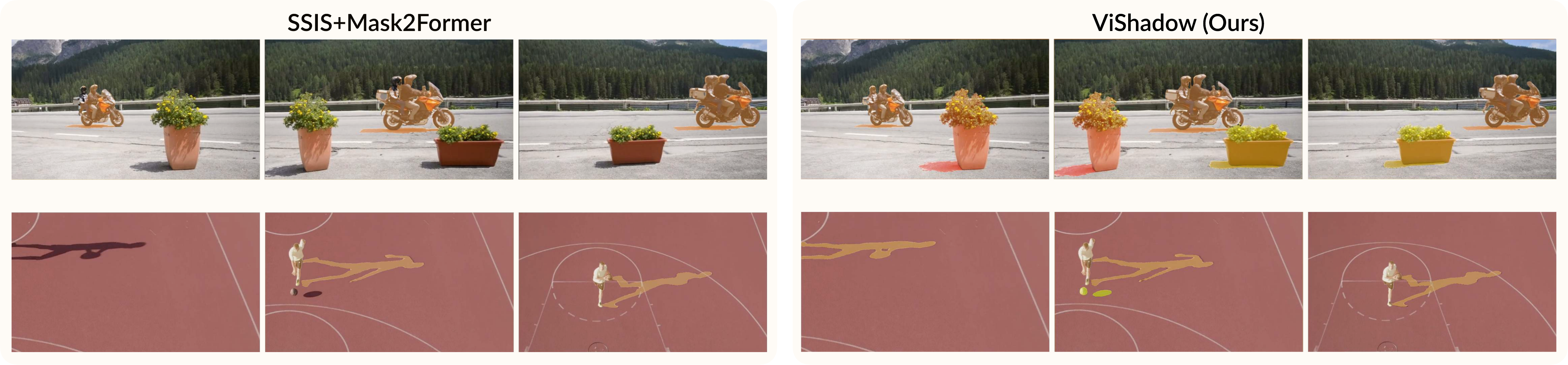}
    \caption{Visual comparison of results produced by our method and SSIS+Mask2former on typical scenarios. Each row displays three frames from a video clip. Detected shadows and objects that are associated are marked in the same color. The combination of SSIS and Mask2Former is limited in its ability to track undefined-category objects and out-of-view object/shadow.}
    \vspace*{-3mm}
    \label{fig:SSIS_Mask2Former_comparision}
\end{figure*}

The surge in internet video content has heightened the demand for video generation and advanced video editing capabilities. In this context, we introduce the \textbf{Video Instance Shadow Detection} (VISD) task, which encompasses \textit{not only the identification of shadows and their associated objects in video frames but also the continuous tracking of each shadow, object, and their associations throughout the entire video sequence.}
Extending the instance shadow detection task to videos presents two primary challenges: \textit{(i) the labor-intensive process of annotating diverse video data, involving the labeling for the detection and tracking of individual shadows and objects along with their temporal associations, and (ii) 
previous works~\cite{wang2020instance,wang2021single,wang2023instance} solely focus on detecting shadow-object pairs and overlook the individual shadows or objects (as shown in the bottom of  Fig.~\ref{fig:SSIS_Mask2Former_comparision}). Consequently, these methods fail to track frames with occlusion or instances outside the field of view. Therefore, it is necessary to retrieve these single parts of pairs in these scenarios.}

To tackle these challenges, we introduce ViShadow, a semi-supervised framework for video instance shadow detection that leverages both labeled image datasets and unlabeled video sequences.
Initially, we use labeled image data~\cite{wang2020instance} to train ViShadow, enabling it to identify shadow instances, object instances, and their spatial relations. Subsequently, we employ center contrastive learning~\cite{he2020momentum} to pair shadows/objects across different images.
In the next phase, we leverage unlabeled videos, introducing an associated cycle consistency loss to enhance tracking ability by exploiting spatial relations and temporal information between shadow and object instances across frames.
Lastly, we tackle the issue of temporary disappearance in videos, especially in occluded scenarios, through a retrieval mechanism, which enhances the tracking of shadow or object instances when portions of them vanish in specific frames. It is achieved by matching the learned tracking embedding for each shadow or object instance across various video frames, thereby improving the overall continuity of shadow-object associations throughout the video.
{Following~\cite{wang2020instance}, we only detect the cast shadows produced by the corresponding objects and ignore the self-shadows on the objects themselves.}
{Also, our method is designed specifically for scenarios involving a single light source, such as those under the sun and sky, and may not be directly applicable to environments with multiple light sources.}

In contrast to our previous works~\cite{wang2020instance,wang2021single,wang2023instance}, this paper not only detects shadows and their associated objects in video frames but also tracks each shadow, object, and their associations continuously throughout the entire video sequence, even when paired shadow or object instances temporarily disappear.
Additionally, we formulate the SOBA-VID dataset, comprising unlabeled videos for training and labeled videos for testing. Accompanying this, the SOAP-VID metric, an extension of the SOAP metric~\cite{wang2020instance}, facilitates the quantitative evaluation of VISD solutions. This metric, incorporating Intersection-over-Union (IoU) computation in a spatiotemporal context, allows for the assessment of the model's performance in detecting and tracking shadow-object associations in video sequences.

The tracking masks derived from ViShadow, used to predict shadow-object associations, offer a valuable tool for various video editing tasks.
For instance, as depicted in Fig.~\ref{fig:teaser}, our framework enables the seamless extraction and insertion of a shadow-object pair from one video to another (or the same video). Additionally, VISD supports video inpainting~\cite{li2022towards,liu2021fuseformer} and shadow editing, allowing for the obfuscation of entire objects by simultaneously addressing the object and its associated shadow, as illustrated in Figure~\ref{fig:inpainting}.
Moreover, leveraging the capabilities of current large vision-language models~\cite{openai2023gpt4,liu2023interngpt}, we achieve text-instructed shadow-object manipulation in videos. This involves interactive editing where user requirements are inputted through text commands, showcasing the potential of our framework for user-guided video editing.
{Our dataset, code, trained model, and evaluation metrics are publicly available at \url{https://github.com/HarryHsing/Video-Instance-Shadow-Detection}.}

Our contributions are summarized as follows:

\begin{itemize}
\item We develop the ViShadow framework for Video Instance Shadow Detection (VISD), leveraging knowledge from both labeled image data and unlabeled video data. This approach enables ViShadow to effectively detect, track, and segment shadow-object associations in videos.

\item We design a retrieval mechanism within ViShadow to handle challenging scenarios such as occlusions or the temporary absence of shadows/objects, ensuring consistent tracking across diverse video situations.

\item  We curate the SOBA-VID dataset, comprising 232 training video sequences and 60 labeled testing videos, along with the SOAP-VID metric for the quantitative evaluation of shadow-object tracking performance in videos.

\item  We integrate ViShadow into a platform capable of mask-guided and text-guided shadow-object editing in videos. It showcases the adaptability of using the results of video instance shadow detection across diverse tasks, including video inpainting, instance cloning, shadow editing, and text-instructed shadow-object manipulation.
\end{itemize}

\vspace*{-2mm}
\section{Related Work }
\label{sec:relatedwork}

\begin{figure*}[tp]
	\centering
	\includegraphics[width=.9\linewidth]{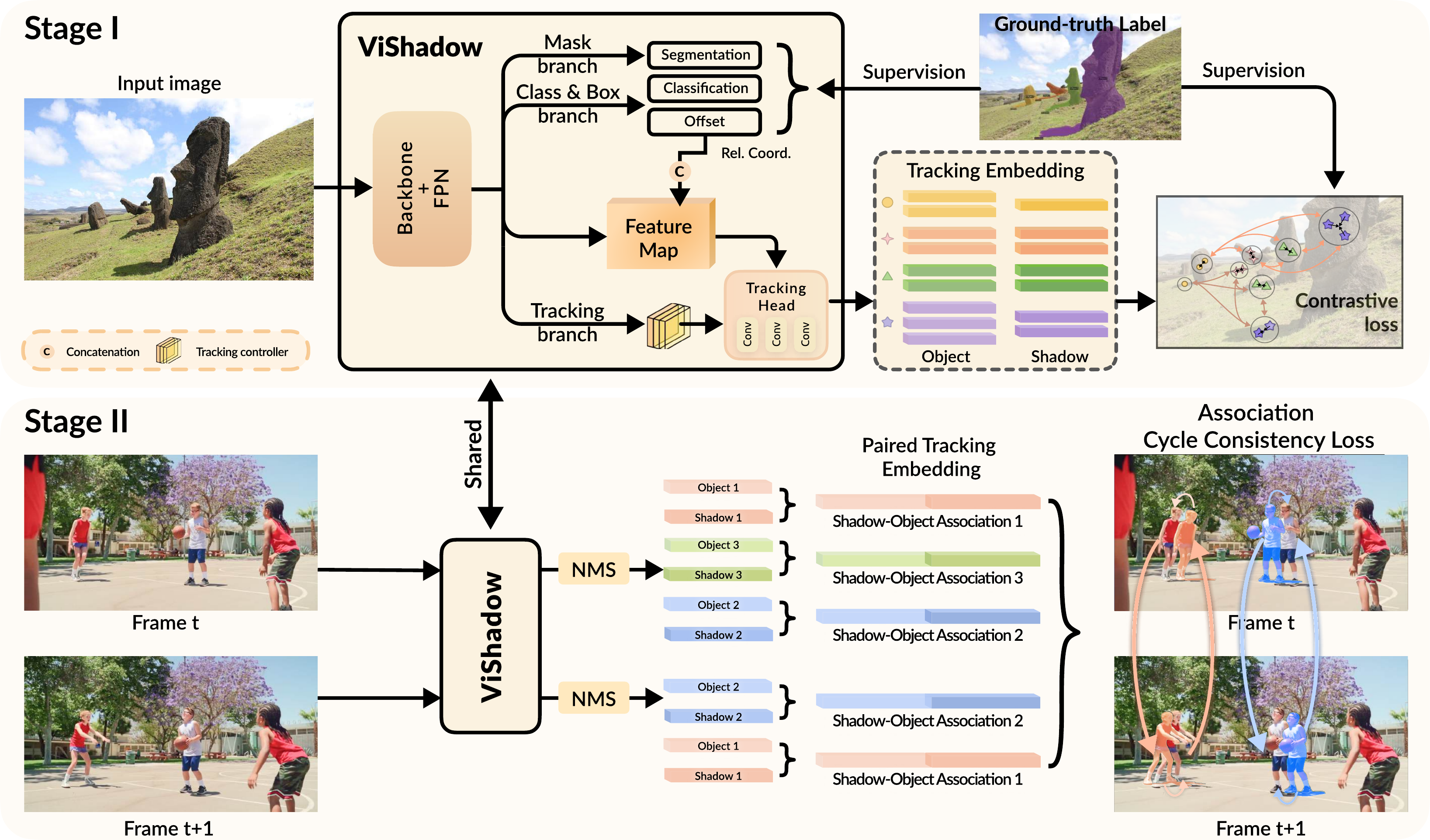}
 \vspace{-2mm}
	\caption{
        The schematic of our semi-supervised video instance shadow detection framework, ViShadow. The top stage involves supervised learning from labeled images, while the bottom stage employs self-supervised learning from unlabeled videos. NMS denotes non-maximum suppression.}
        \vspace{-3mm}
	\label{fig:figure4}
\end{figure*}

\subsection{Shadow Detection}
Image shadow detection estimates a binary shadow region mask using methods ranging from physical techniques~\cite{salvador2004cast, finlayson2005removal, panagopoulos2011illumination, tian2016new}, user input~\cite{arbel2010shadow, zhang2015shadow}, to handcrafted features~\cite{finlayson2009entropy, lalonde2010detecting, zhu2010learning, guo2011single, huang2011characterizes}. Recent advancements in deep learning~\cite{hu2018direction,zhu2018bidirectional, zheng2019distraction, le2018a+d, hu2020direction, chen2020multi, hu2021revisiting, yang2023silt} have demonstrated remarkable performance. Extending this to video, Chen et al.~\cite{chen2021triple} introduced ViSha and TVSD-Net, focusing on intra-video and inter-video feature learning. SC-Cor~\cite{ding2022learning} improves temporal coherence, Lu et al.~\cite{lu2022video} addresses generalization errors, and Liu et al.~\cite{liu2023scotch} handles large shadow deformations. Please see~\cite{hu2024unveiling} for the comprehensive survey on image and video shadow detection with deep learning.
In contrast, our video instance shadow detection edits or removes individual shadows with associated objects at the video level, providing a unique perspective.

\subsection{Instance Shadow Detection}
Instance shadow detection identifies shadows and their associations with objects. LISA~\cite{wang2020instance} pioneers this with a two-stage end-to-end framework for predicting bounding boxes, masks, and associations in images, featuring a strategy for pairing instances. SSIS~\cite{wang2021single,wang2023instance} adopts a single-stage fully convolutional network with bidirectional relation learning for understanding shadow-object relationships. The extension to video instance shadow detection encompasses both instance detection and paired tracking in videos. However, this extension introduces challenges not seen in image-based instance shadow detection, involving labor-intensive video annotation on a per-frame basis and tracking partial shadow-object pairs due the occlusion or out-of-view 
issues in videos, adding complexity to the task. Please see~\cite{hu2024unveiling} for the comprehensive survey and comparisons.

\subsection{Video Instance Segmentation}
Video instance segmentation involves the detection and tracking of individual objects in videos, categorized into online and offline methods. Online methods, referred to as tracking-by-detection, detect instances frame by frame and match them accordingly. Pioneering methods like MaskTrack R-CNN~\cite{yang2019video} integrate tracking heads to predict embeddings for matching instances. Offline methods directly segment instances in videos, treating each as a sub-volume in the spatio-temporal volume. Approaches like VisTR~\cite{wang2021end}, IFC~\cite{hwang2021video}, SeqFormer~\cite{wu2022seqformer}, and Mask2Former~\cite{cheng2021mask2former} treat video instance segmentation as a sequence decoding problem. A recent semi-supervised approach~\cite{fu2021learning} does not rely on video annotations. 
Existing approaches for video instance segmentation are often limited to specific object categories and typically require rich features from objects for effective tracking, posing a challenge given the limited features of shadow instances. 
In contrast, our proposed video instance shadow detection approach requires no reliance on video instance annotations, concentrating on detecting and tracking foreground objects along with their associated shadows.
Most importantly, even the shadow or object instance is disappeared in several frames, our method is still to detect and track the left part of shadow-object association, as the comparisons shown in Fig.~\ref{fig:SSIS_Mask2Former_comparision}.

\section{SEMI-Supervised Video Instance Shadow Detection Framework}
\label{sec:method}

\textit{Video Instance Shadow Detection} (VISD) encompasses detecting, segmenting, associating, and tracking shadow-object pairs in videos. Addressing challenges such as labor-intensive video data annotation and complexities from occlusion, we introduce a semi-supervised framework (ViShadow) leveraging knowledge from labeled images and unlabeled videos, featuring a bidirectional retrieval mechanism for tracking unpaired shadow/object instances.

\subsection{Overall Network Architecture}
The overall architecture of the semi-supervised video instance shadow detection framework (ViShadow) is depicted in Fig.~\ref{fig:figure4}, featuring two stages:
(i) Learning from labeled images: Drawing upon labeled images from SOBA~\cite{wang2021single}, as illustrated in Stage I, we employ a Convolutional Neural Network (CNN) to extract feature maps at multiple resolutions, enhanced by a Feature Pyramid Network (FPN)~\cite{lin2017feature}. Subsequently, we utilize multiple branches on these feature maps: a mask branch for predicting shadow and object instance masks, and a class \& box branch for identifying object, shadow, and background pixels. The class \& box branch also predicts the offset, indicating the relationships between shadow and object instances.
These two branches are designed to learn the detection and segmentation of shadow/object instances, as well as the associations between shadows and objects, guided by image-level supervision.
To enable instance-paired tracking, we introduce the tracking branch, which comprises a dynamic tracking controller and a tracking head. The tracking controller dynamically generates network parameters (filters) assigned to the tracking head. Subsequently, the tracking head applies convolutions to the feature map using filters obtained from the tracking tower based on the detected instance. The resulting output of the tracking head is comprised of tracking embeddings.
Subsequently, we employ a center contrastive loss to train the tracking embeddings by discerning similarities or differences between shadow/object instances.
(ii) Learning from unlabeled videos: Harnessing unlabeled videos, we employ temporal correspondence to augment the paired tracking capability of the tracking branch, which is trained in a self-supervised manner.
As shown in Stage II, we adopt ViShadow trained in Stage I to produce tracking embeddings for instances in the neighboring frames. Through our introduced association cycle consistency loss, we regularize the embeddings of identical shadow-object associations to be more proximate.

In the following subsections, we elaborate on the methodology for acquiring a tracking embedding for instance-paired tracking (Section~\ref{Sec:method-self}). Additionally, we introduce a bidirectional retrieving mechanism aimed at pairing up shadows or objects that have temporarily disappeared (Section~\ref{Sec:method-bidirectional}).

\begin{figure}[!t]
    \centering
    \includegraphics[width=1\linewidth]{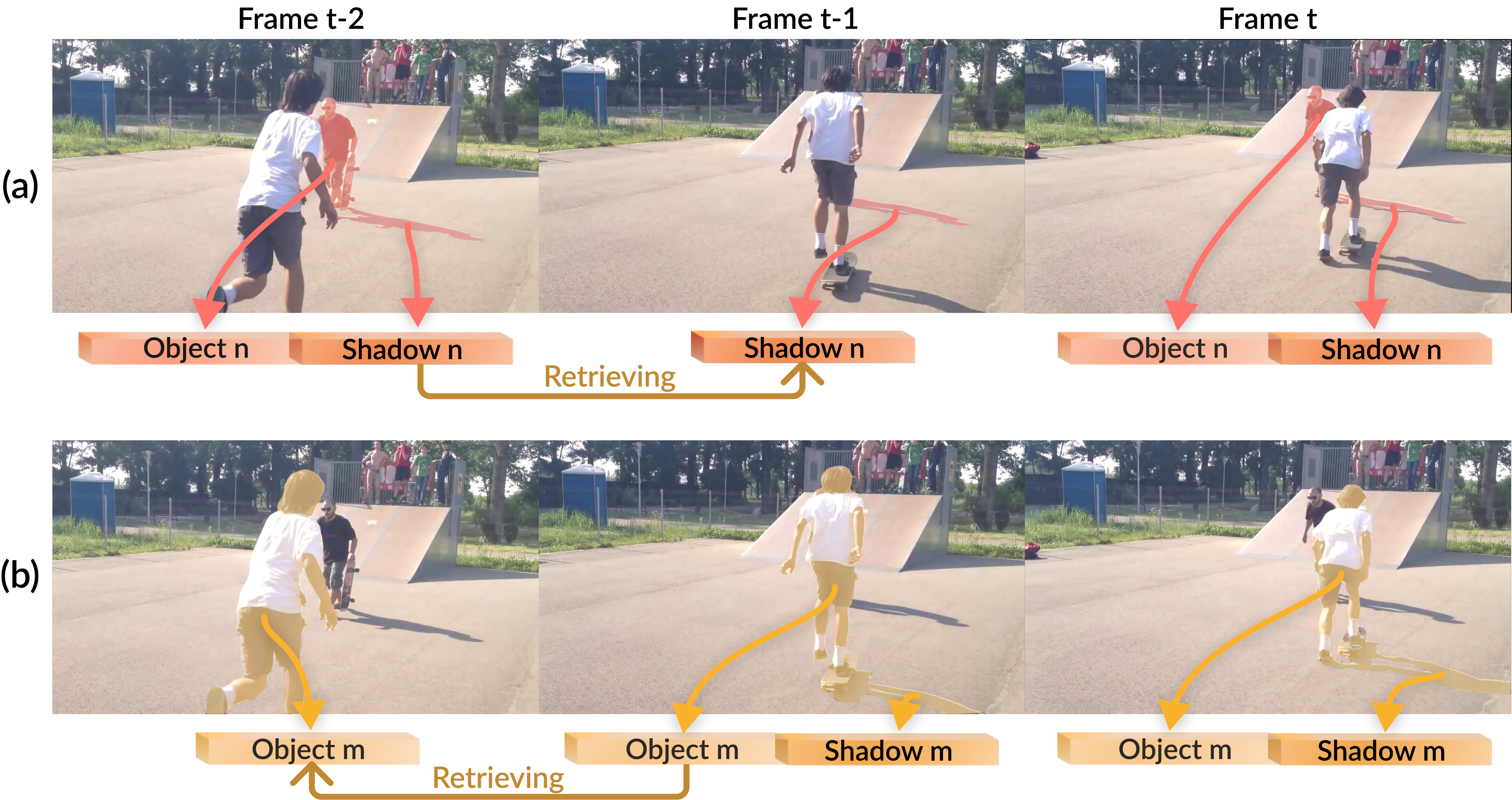}
    \caption{The schematic illustration of the proposed bidirectional retrieving mechanism. }
    \label{fig:retrieve}
\end{figure}

\subsection{Instance Paired Tracking}
\label{Sec:method-self}
%
%

\subsubsection{Learning from Labeled Images}

In this section, we outline the training methodology for ViShadow using labeled image data, as depicted in Stage I of Fig.~\ref{fig:figure4}.
For a single image, we employ mask and class \& box mask branches~\cite{wang2021single} to detect and segment shadow and object instances. To enhance tracking capabilities, we introduce a tracking branch with a dynamic tracking controller. This controller generates parameters for convolutional filters, used by the tracking head to predict instance tracking embeddings, taking feature maps and relative coordinates as inputs.

To learn tracking ability from image-level supervisions, we propose a center contrastive learning approach~\cite{he2020momentum}. We aim to bring tracking embeddings of the same instance closer and maintain distinctiveness from other instances.
For each instance $\Omega {i}$, the center embedding $C{i}$ is computed as the average of all instance-level tracking embeddings:
\begin{equation}
C_{i} \ = \ \frac{1}{N_{i} } {\textstyle \sum_{e\in \Omega {i} }^{}} f{e} \ ,
\end{equation}
with $N_{i}$ denoting the number of locations on the feature maps belonging to $\Omega _{i}$ and a classification score $\textgreater$ 0.05~\cite{tian2019fcos}.

To encourage closer embeddings of the same instance, the center loss minimizes the L1 distance:
\begin{equation}
\label{equtation:eq1}
L_{i}^{center} \ = \ \sum_{e\in \Omega {i} }^{} \parallel C{i} - f_{e} \parallel \ .
\end{equation}
Ensuring distinctiveness from other instances, we compute a similarity matrix $Sim(i,j)$ for shadows and objects separately. The contrast loss, based on the cross entropy $C_{en}$, maximizes self-matching likelihood and separates different instances:
\begin{equation}
\label{equtation:eq3}
L^{contra} \ = \ C_{en}(Sim_{S},I) + C_{en}(Sim_{O},I) \ ,
\end{equation}
with $I$ as an identity matrix.
Joint optimization of the framework with $L_{i}^{center}$ and $L^{contra}$ facilitates effective learning of tracking abilities from input images.

\begin{figure*}[tp]
	\centering
	\includegraphics[width=1\linewidth]{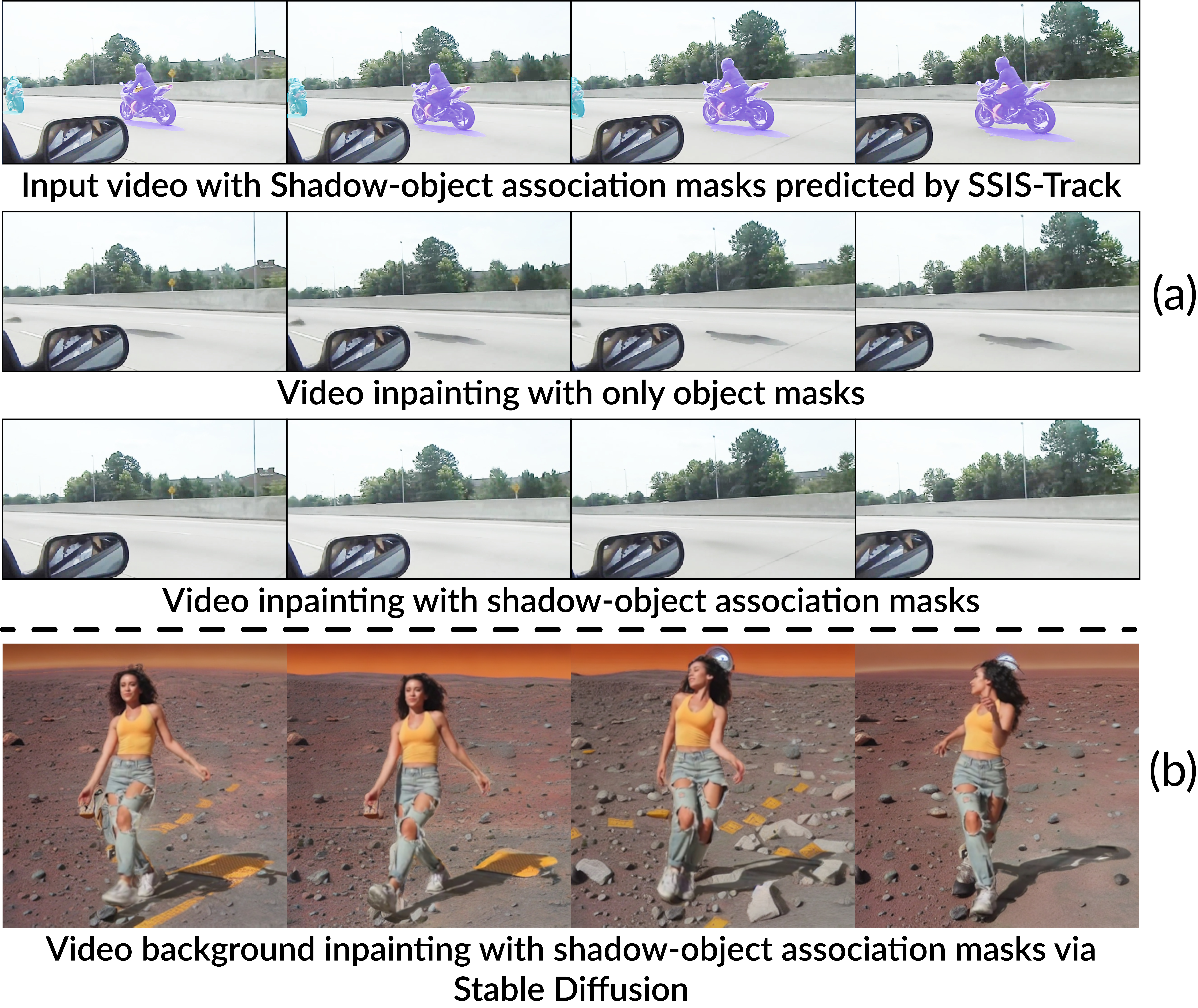}
	\caption{Example sequences demonstrate the application of video instance shadow detection in video inpainting.
        }
	\label{fig:inpainting}
\end{figure*}

\subsubsection{Learning from Unlabeled Videos}


In contrast to labeled image data providing instance labels, we utilize unlabeled videos for self-supervised learning in our ViShadow framework, as illustrated in Fig.~\ref{fig:figure4} Stage II. {Note that the center contrastive learning is not employed in Stage II due to the difficulty of accurately forming positive and negative pairs across frames in the absence of ground truth labels.}

For two consecutive frames in a video, we compute tracking embeddings 
of detected instances.
Non-maximum suppression (NMS) is then applied, aiming to obtain distinct paired shadow-object tracking embeddings for each instance.
Subsequently, a transition matrix $A$ is generated by computing the embedding similarity of instance pairs at different time frames. 
Each element $A_{t}^{t+1}(i,j)$ represents the transition probability from the $i^{th}$ instance at time $t$ to the $j^{th}$ instance at time $t+1$.

Motivated by~\cite{jabri2020space, li2019joint, wang2019learning}, we introduce an associated cycle consistency loss. If the $i^{th}$ instance at time $t$ can transition (similarly) to the $j^{th}$ instance at time $t+1$, the inverse transition should also hold. 
We encourage large values in the diagonal elements of the multiplication results of the transition matrices $A_{t}^{t+1}$ and $A_{t+1}^{t}$ by comparing them with an identity matrix ($I$):
\begin{equation}
\label{equtation:eq4}
L^{cyc} \ = \ C_{en}(A_{t}^{t+1}A_{t+1}^{t},I) \ .
\end{equation}

It is worth noting that our associated cycle consistency loss utilizes paired shadow-object tracking embeddings and is conducted bidirectionally.

\subsubsection{Tracking Paired Instances Across Frames.} 
To facilitate instance paired tracking, we employ a matching strategy inspired by~\cite{yang2019video} to follow each paired shadow-object association across consecutive frames.
Initially, we assume the existence of $N$ instances in the tracking queue, each characterized by their tracking embeddings represented through instance features. 
Subsequently, we consider the appearance of $M$ instances in the subsequent video frame, computing their corresponding tracking embeddings.

Furthermore, we calculate the similarity $Score(f_i, f_j)$ between instances in the current frame and those in the tracking queue by evaluating the cosine similarity of their tracking embeddings. The resulting similarity is normalized through a bidirectional SoftMax function:

\begin{equation} 
	\label{equtation:eq5}
	\begin{split} 
		Score(f_i,f_j) = &\frac{0.5 \exp(\cos(f_{i}, f_{j}))}{\sum_{k=1}^{M} \exp(\cos(f_{k}, f_{j}))}\\
		& + \frac{0.5 \exp(\cos(f_{i}, f_{j}))}{\sum_{k=1}^{N} \exp(\cos(f_{i}, f_{k}))} \ , 
	\end{split} 
\end{equation}

where $f_{i}$ denotes the tracking embedding of the detected instance in the current frame, and $f_{j}$ denotes the latest instance tracking embedding in the tracking queue.

If the score falls below a predefined matching threshold, we consider the instance as a new identity object and add its embedding to the tracking queue. Conversely, if the score exceeds the threshold, we regard it as a tracked instance and update the corresponding embedding in the tracking queue.
It is noteworthy that in this strategy, we concatenate the tracking embeddings of paired shadow-object associations to form the overall tracking embedding.

\begin{table*}[tp]
    \centering
        \centering
        \caption{Evaluation on the ViShadow framework design.}
        \label{table:2}
            \begin{tabular}{@{}l|c|ccc|c|c|c@{}}
            \toprule
                Method & \makecell{With \\ Tracking \\ Branch} & \makecell{Contrastive \\ Learning} & \makecell{Temporal \\ Correspondence} & \makecell{Retrieving \\ Mechanism} & \makecell{SOAP-\\VID} & \makecell{Association \\ AP} & \makecell{Instance \\ AP} \\
                \midrule
                SSIS & & & & & 29.7 & 40.0 & 36.5 \\
                \midrule
                \multirow{3}{*}{ViShadow} & $\checkmark$ & $\checkmark$ & & & 36.8 & 55.4 & 45.2 \\
                 & $\checkmark$ & $\checkmark$ & $\checkmark$ & & 38.0 & 55.9 & 46.8 \\
                 & $\checkmark$ & $\checkmark$ & $\checkmark$ & $\checkmark$ & \textbf{39.6} & \textbf{61.5} & \textbf{50.9} \\
            \bottomrule
            \end{tabular}

\end{table*}

\subsection{Bidirectional Retrieving Mechanism for Unpaired Shadow/Object Instances}
\label{Sec:method-bidirectional}

In the domain of video instance shadow detection, instances of shadows or objects may encounter brief interruptions, whether caused by temporary occlusion or displacement beyond the camera's field of view. 
To address this, our bidirectional retrieving mechanism, employed solely during inference, facilitates tracking individual shadow/object instances and locating their associated object/shadow instances upon reappearance in the video. 

In the matching process, we adopt Eq.~\eqref{equtation:eq5} to compute the similarity matrix, comparing the tracking embeddings of individual shadow/object instances in the current frame with those in the tracking queue. 
A successful match prompts an update to the tracking embedding of the individual shadow/object instance in the queue, while preserving the corresponding tracking embedding of the paired object/shadow instance. This enables the tracking of the paired shadow-object association upon its reappearance. 
Instances with unmatched tracking embeddings are disregarded, as our system exclusively handles paired shadow-object associations that persist for at least one video frame. 
To address situations where paired shadow-object associations occur in subsequent frames, the mechanism is applied once more in reverse order.

Fig.~\ref{fig:retrieve} illustrates two scenarios: (i) In (a), object $n$ disappears in time frame $t-1$, but its shadow persists. 
Our system successfully tracks the shadow instance and re-establishes its association with the object in time frame $t$; 
(ii) In (b), shadow $m$ is visible only in time frames $t-1$ and $t$. 
Using our retrieving mechanism in reverse, we manage to associate object $m$ with its previous appearance in time frame $t-2$. 
Leveraging tracking embeddings, we can reclaim unpaired instances, enhancing performance and yielding improved visual effects in video instance shadow detection applications.

{During the training phase, our framework uses an association cycle consistency loss to ensure the consistency of tracking embeddings across frames. This loss helps maintain the identity of shadow-object pairs even when they temporarily disappear. In the testing phase, the bidirectional retrieval mechanism becomes crucial. If an instance (shadow or object) disappears in a frame, the mechanism stores its last known embedding. When the instance reappears, the retrieval mechanism uses the stored embeddings to re-associate the shadow and object, maintaining the continuity of their association. This approach effectively handles occlusions and temporary disappearances, ensuring robust tracking of shadow-object pairs throughout the video sequence.}

\section{Dataset and Evaluation Metrics}

\begin{figure}[tp]
    \centering
    \includegraphics[width=1\linewidth]{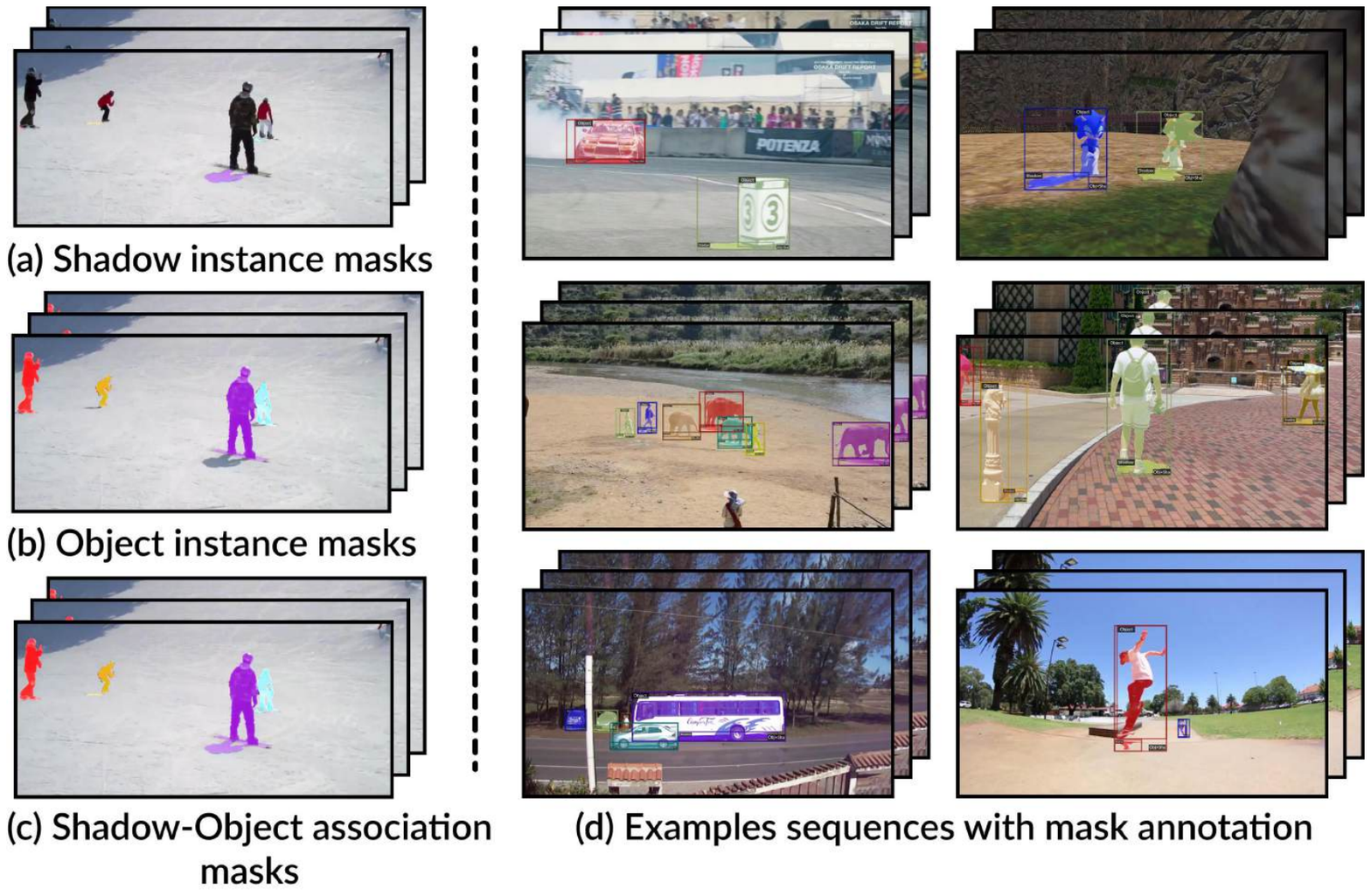}
    \caption{Example frames in our SOBA-VID dataset contain
    (a) shadow instances,
    (b) object instances, and
    (c) their associations.
    More examples are presented in (d).
    }
    \label{fig:figure2}
\end{figure}

\subsection{Dataset}
\label{sec:dataset}
We introduce SOBA-VID (Shadow-Object Association of Videos), a novel dataset tailored for video instance shadow detection. In its compilation, we adhere to specific criteria:
(i) Ensuring the presence of common shadow-object pairs in the collected videos.
(ii) Ensuring the clarity and absence of ambiguity in the shadows depicted in the videos.
(iii) Ensuring diversity in instance categories and scenarios, encompassing both occluded shadows and objects.
(iv) Inclusion of videos with varied backgrounds in motion and diverse light directions.

SOBA-VID comprises 292 videos totaling 7045 frames, sourced from (i) existing datasets, including DAVIS~\cite{Caelles_arXiv_2018}, YouTube-VIS~\cite{yang2019video}, and OVIS~\cite{qi2022occluded}, (ii) self-collected videos, and (iii) Internet searches using the keyword ``shadow plus common instances.'' 
We randomly partition the videos into a training set (232 videos, 5863 frames) and a testing set (60 videos, 1182 frames). 
For the test set videos, we meticulously annotate frame-by-frame masks of each shadow/object instance using Apple Pencil, forming a total of 134 shadow-object pairs (refer to Fig.~\ref{fig:figure2} for examples).
{For the training set videos, we annotate one frame in every four frames, while ensuring that the last frame of each video is always annotated. This results in a total of 1719 annotated frames, containing 503 shadow-object pairs. Note that our semi-supervised framework does not use these annotations in Stage II, and we hope the labeled frames will promote future research in this field.}
{Overall, our dataset contains 292 videos with a total of 7045 frames and 637 annotated shadow-object pairs. The average video length is 24.1 frames, and there are on average 2.2 shadow-object pairs per video.}


\subsection{Evaluation Metrics}
\label{sec:metrics}
For the assessment of video instance shadow detection, we extend the evaluation metric SOAP~\cite{wang2020instance} originally designed for single-image instance shadow detection to accommodate video sequences. 
A sample is deemed a true positive if the Intersection-over-Union (IoU) between the predicted and ground-truth shadow instances, object instances, and shadow-object associations all surpass a specified threshold $\tau$. 
To incorporate the temporal aspect, we replace the IoU in SOAP with the spatio-temporal IoU as proposed in~\cite{yang2019video}, resulting in the updated metric named SOAP-VID. 
The reported performance is averaged over multiple $\tau$ values in the range [0.5:0.05:0.95]. 
Additionally, utilizing spatio-temporal IoU, we compute the average precision for both shadow/object instances and paired shadow-object associations across thresholds [0.5:0.05:0.95], denoted as Instance AP and Association AP, respectively.

\begin{figure*}[tp]
    \centering
\includegraphics[width=1\linewidth]{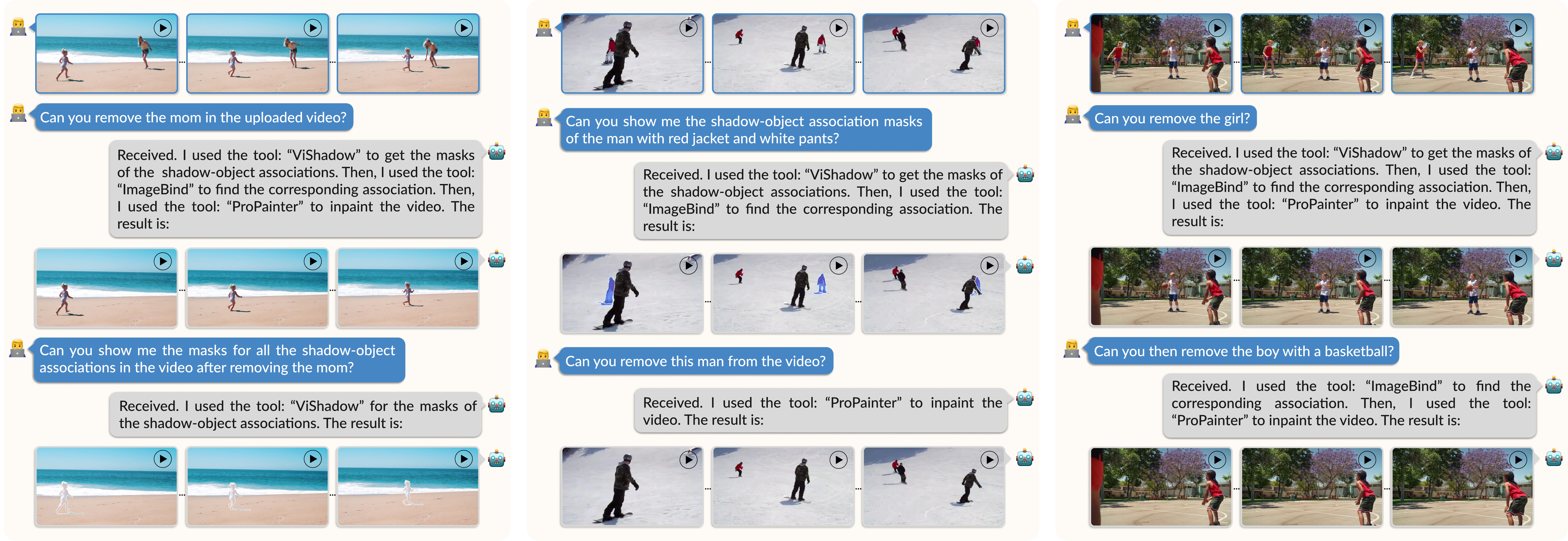}
    \caption{Text-instructed video editing platform integrated with ViShadow, which supports shadow-object  edition in videos. Please watch the demo video in the supplementary material.}
    \label{fig:text_visd}
\end{figure*}

\begin{table}[tp]
\centering
    \caption{Comparison with three baseline methods for video instance shadow detection, showcasing our method's substantial superiority. Improvements over ``SSISv2 + Mask2Former'' on SOAP-VID, Association AP, and Instance AP are 27.7\%, 34.6\%, and 31.2\%, respectively.  }
\label{table:1}
\resizebox{\linewidth}{!}{%
\setlength{\tabcolsep}{1mm}
\begin{tabular}{@{}l|@{\hspace{1mm}}c@{\hspace{1mm}}c@{\hspace{1mm}}c@{}}
\toprule
Method                   & \multicolumn{1}{l}{SOAP-VID} & \multicolumn{1}{l}{Association\,AP} & \multicolumn{1}{l}{Instance\,AP} \\ \midrule
SSIS~\cite{wang2021single} + IoU Tracker\cite{bochinski2017high}         & 21.5                               & 31.7                                        & 25.7                              \\
SSIS~\cite{wang2021single} + Mask2Former\cite{cheng2021mask2former}  & 29.2                                & 42.5                                        & 36.2                             \\
SSISv2~\cite{wang2023instance} + Mask2Former\cite{cheng2021mask2former}  & {31.0}                                & {45.7} & {38.8}                             \\
\textbf{ViShadow} & \textbf{39.6}                     & \textbf{61.5}                               & \textbf{50.9}                               \\ \bottomrule
\end{tabular} }
\end{table}

\section{Experimental Results}
\label{sec:experiments}

\subsection{Implementation Details}

We follow the training protocols of CondInst~\cite{tian2020conditional} and AdelaiDet~\cite{tian2019adelaidet} for our framework. For image training, the backbone network's weights are initialized with those from CondInst trained on COCO~\cite{lin2014microsoft}. We then utilize the same parameters in SSIS~\cite{wang2021single,wang2023instance} to train ViShadow on the SOBA dataset~\cite{wang2020instance}. When transitioning to video training, we use the weights of ViShadow trained on images as initialization, set the mini-batch size to four, and optimize ViShadow on an NVIDIA RTX 3090 GPU for 5,000 iterations. The initial learning rate starts at $1e-5$ and is reduced to $1e-6$ at 4,500 iterations. Additionally, we re-scale input images without altering the aspect ratio, ensuring a minimum shorter side length of 672 pixels.

%
For inference, ViShadow processes input videos online. Its tracking branch predicts tracking embeddings for both paired and unpaired shadow/object instances on a frame-by-frame basis, employing the matching strategy outlined in Section~\ref{Sec:method-self}. Notably, ViShadow demonstrates its efficiency by completing a 21-frame video in 18 seconds.

\subsection{Comparison with Baseline Methods}
We compare our method with three baselines: (i) utilizing image-level detector SSIS~\cite{wang2021single} to detect the shadow instances, object instances, and shadow-object associations on each frame, and employing the IoU Tracker \cite{bochinski2017high} to match instances across frames based on IoU values between detected bounding boxes; and (ii) 
employing the tracking algorithm Mask2former~\cite{cheng2021mask2former} to detect and track object instances across all video frames, using SSIS~\cite{wang2021single} for shadow/object instance and shadow-object association detection on each frame, merging the results based on mask IoU of each object instance, and similarly using SSISv2~\cite{wang2023instance} for the same task and merging the results in the same manner.

Table~\ref{table:1}\footnote{The hyperparameters for the post-processing in SSIS + Mask2Former and SSISv2 + Mask2Former have been unified for comparison, leading to different results from the previous arXiv version.} presents the comparison results, showing our method significantly outperforming both baselines across all evaluation metrics on the SOBA-VID test set. The IoU Tracker heavily relies on the quality of detection results and frame rate, while Mask2Former is constrained to recognizing objects of the same category as those in the training data~\cite{yang2019video}. In contrast, our approach benefits from end-to-end training and knowledge acquired from unlabeled videos. Visual comparisons between ViShadow and ``SSIS + Mask2Former'' are provided in Fig.~\ref{fig:SSIS_Mask2Former_comparision}.

\begin{figure}[tp]
	\centering
	\includegraphics[width=1\linewidth]{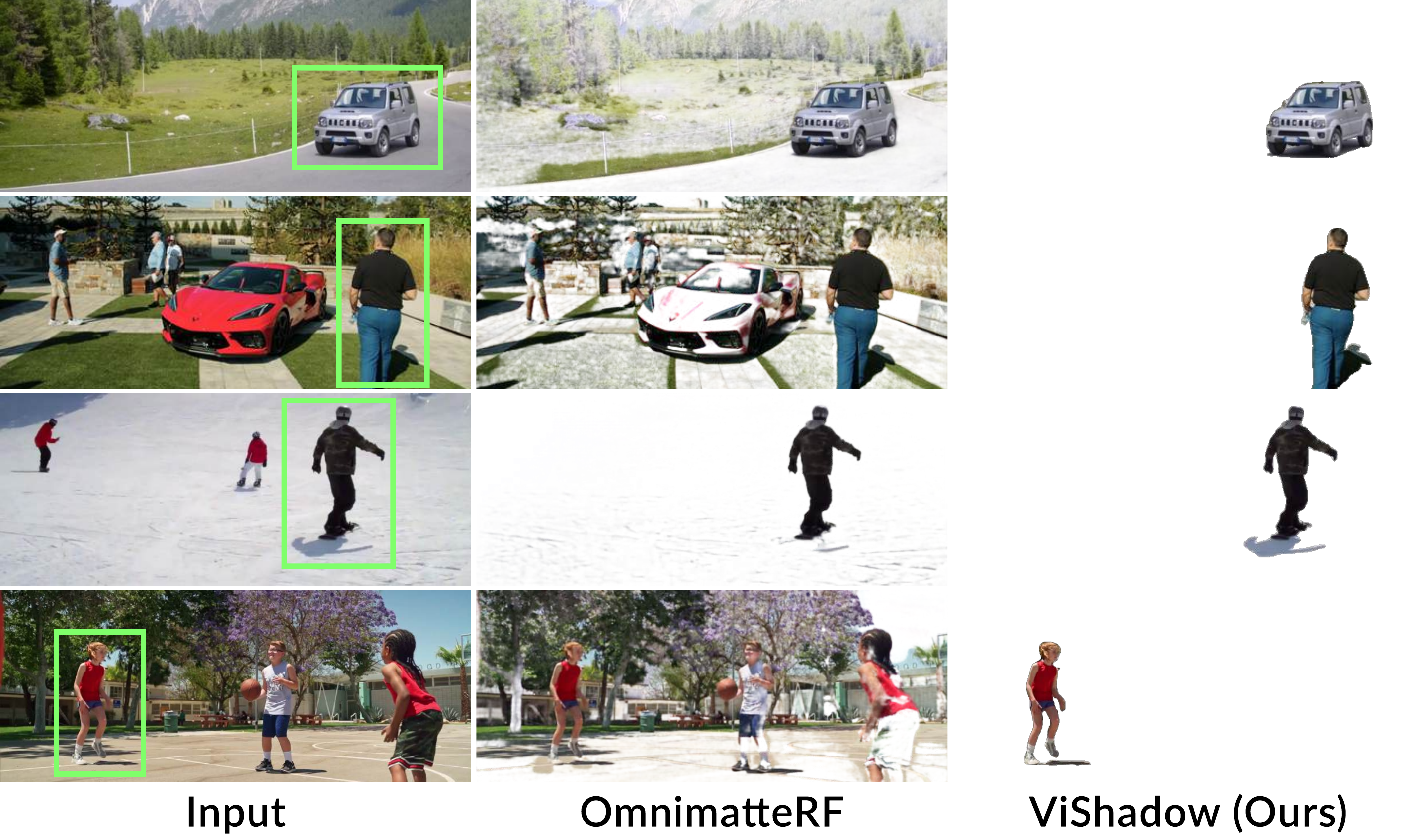}
	\caption{Visual comparisons between the results of OmnimatteRF and our ViShadow.}
	\label{fig:comp_omnirf}
\end{figure}

\begin{table}[tp]
       \centering
        \caption{Evaluation on the instance paired tracking design.}
        \label{table:3}
            \begin{tabular}{@{}l|cc|c|c|c@{}}
            \toprule
                Method & \makecell{Object \\ Embedding} & \makecell{Shadow \\ Embedding} & \makecell{SOAP-\\VID} & \makecell{Asso. \\ AP} & \makecell{Inst. \\ AP} \\
                \midrule
                \multirow{3}{*}{ViShadow} & $\checkmark$ & & 38.4 & 59.7 & 49.3 \\
                 & & $\checkmark$ & 38.3 & 59.1 & 48.6 \\
                 & $\checkmark$ & $\checkmark$ & \textbf{39.6} & \textbf{61.5} & \textbf{50.9} \\
            \bottomrule
            \end{tabular}
\end{table}

\subsection{Ablation Study}

\subsubsection{Component Analysis} 
We conduct an ablation study, incrementally introducing major components—tracking branch with contrastive learning in Eq.\eqref{equtation:eq3}, temporal correspondence in Eq.\eqref{equtation:eq4}, and bidirectional retrieving mechanism—starting from the SSIS baseline, which employs instance tracking through the mask controller's output. The results in Table~\ref{table:2} demonstrate the contributions of each component, with contrastive learning and temporal correspondence enhancing paired tracking, and the retrieving mechanism addressing missed unpaired instances.

\subsubsection{Instance Paired Tracking Analysis} We assess the effectiveness of instance paired tracking. Table~\ref{table:3} indicates that using tracking embeddings for individual shadow/object instances achieves paired tracking. And the best performance across all evaluation metrics is achieved by concatenating both shadow and object embeddings.

\subsection{Comparison with OmnimatteRF}
OmnimatteRF~\cite{lin2023omnimatterf} is designed for creating video mattes, encompassing the primary object and associated elements like reflections and shadows. This innovative approach builds upon the foundational work presented in~\cite{lu2021omnimatte}, seamlessly integrating dynamic 2D foreground layers with a 3D background model. The process of generating video mattes with OmnimatteRF involves extensive data preprocessing, utilizing RAFT~\cite{teed2020raft} for optical flow, MiDaS~\cite{ranftl2020towards} for generating monocular depth maps, and the pose estimation method of RoDynRF~\cite{liu2023robust}. Additionally, user-provided coarse masks of the target object in each frame are required.

In contrast, ViShadow stands out by eliminating the need for complex data preprocessing. Figure~\ref{fig:comp_omnirf} visually compares the results of OmnimatteRF and ViShadow. Notably, within the green boxes highlighting objects, OmnimatteRF exhibits unintended background regions and may not fully capture shadow areas. Furthermore, it struggles with processing static objects due to its reliance on optical flow.
Additionally, OmnimatteRF takes about four hours to process a 21-frame video on a single RTX3090 GPU, attributed to its need for test-time training. 

\section{Applications}
\label{sec:applications}

\subsection{Mask-Guided Shadow-Object Manipulation}

The tracking masks from ViShadow predict shadow-object pairs, offering a powerful tool for various video editing tasks, including inpainting, instance cloning, and shadow editing.

\subsubsection{Video Inpainting} The process of video instance shadow detection involves generating masks that associate shadows with objects across frames. This capability allows for tasks such as object removal with their accompanying shadows or seamless background replacement, all while preserving the integrity of both objects and shadows. In Figure~\ref{fig:inpainting} (left), we showcase the application of ViShadow in video inpainting~\cite{li2022towards}, successfully eliminating motorcycles and their shadows simultaneously. Furthermore, Figure~\ref{fig:inpainting} (right) illustrates the synergy between our masks and Stable Diffusion~\cite{Rombach_2022_CVPR}, demonstrating background replacement by placing a dancer on Mars along with her associated shadow.

\subsubsection{Instance Cloning} 
\begin{figure}[tp]
	\centering
	\includegraphics[width=1\linewidth]{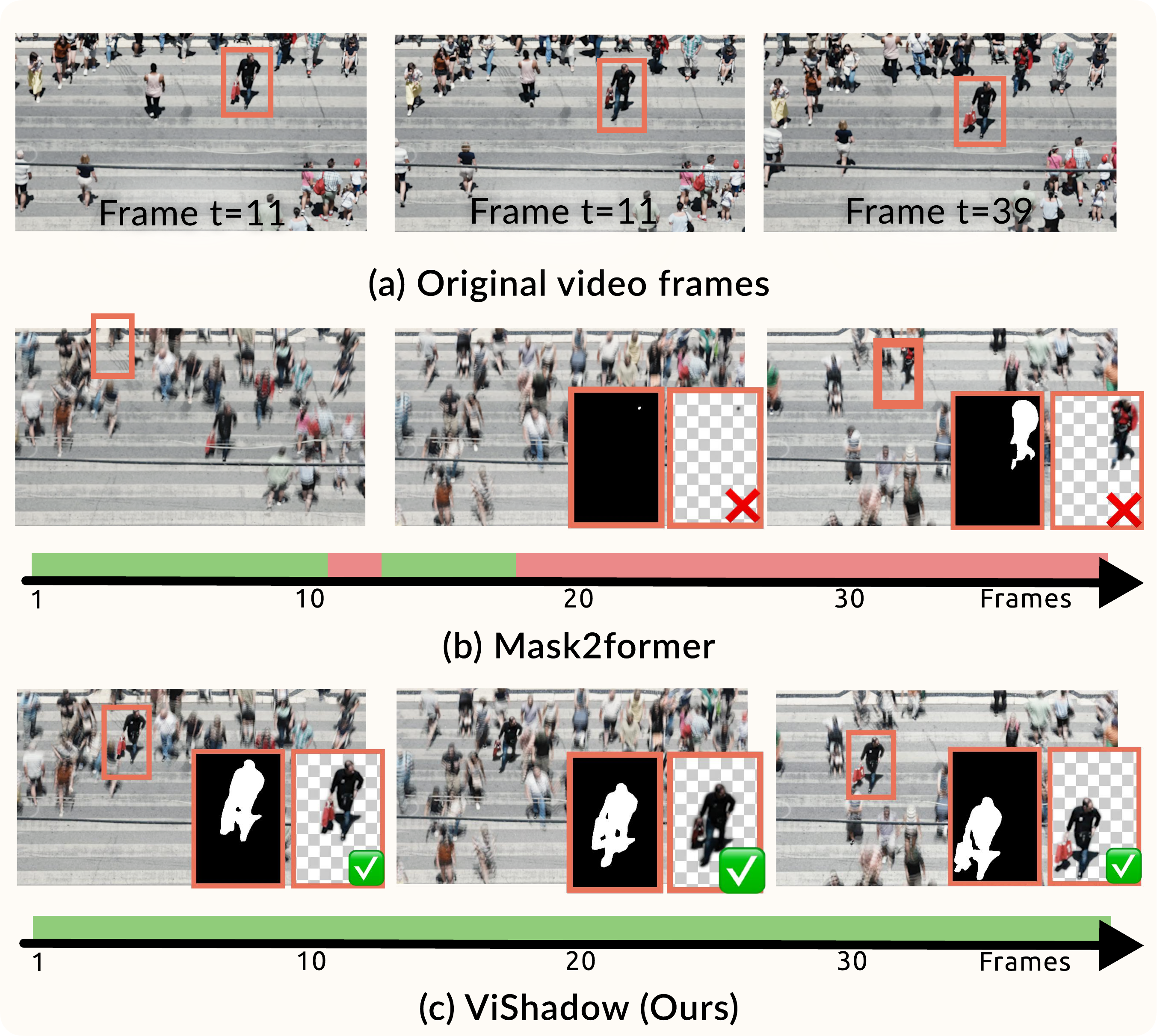}
	\caption{A man walks in a crowd created using our results.}
	\label{fig:mask_visd_1}

\end{figure}

Figure~\ref{fig:mask_visd_1} demonstrates video instance cloning facilitated by video instance shadow detection, enabling the creation of cinematic effects such as a man with a suitcase walking in a crowd flow. Paired shadow-object associations are duplicated across frames from one sequence to another, with adjustments made to frame rates and motion blur. The final result is depicted in Figure~\ref{fig:mask_visd_1} (c), while Figure~\ref{fig:mask_visd_1} (b) contrasts outcomes using Mask2Former\cite{cheng2021mask2former}, which fails to detect instances and produces unrealistic effects without associated shadows.

\subsubsection{Shadow Editing} 
\begin{figure}[tp]
	\centering
	\includegraphics[width=1\linewidth]{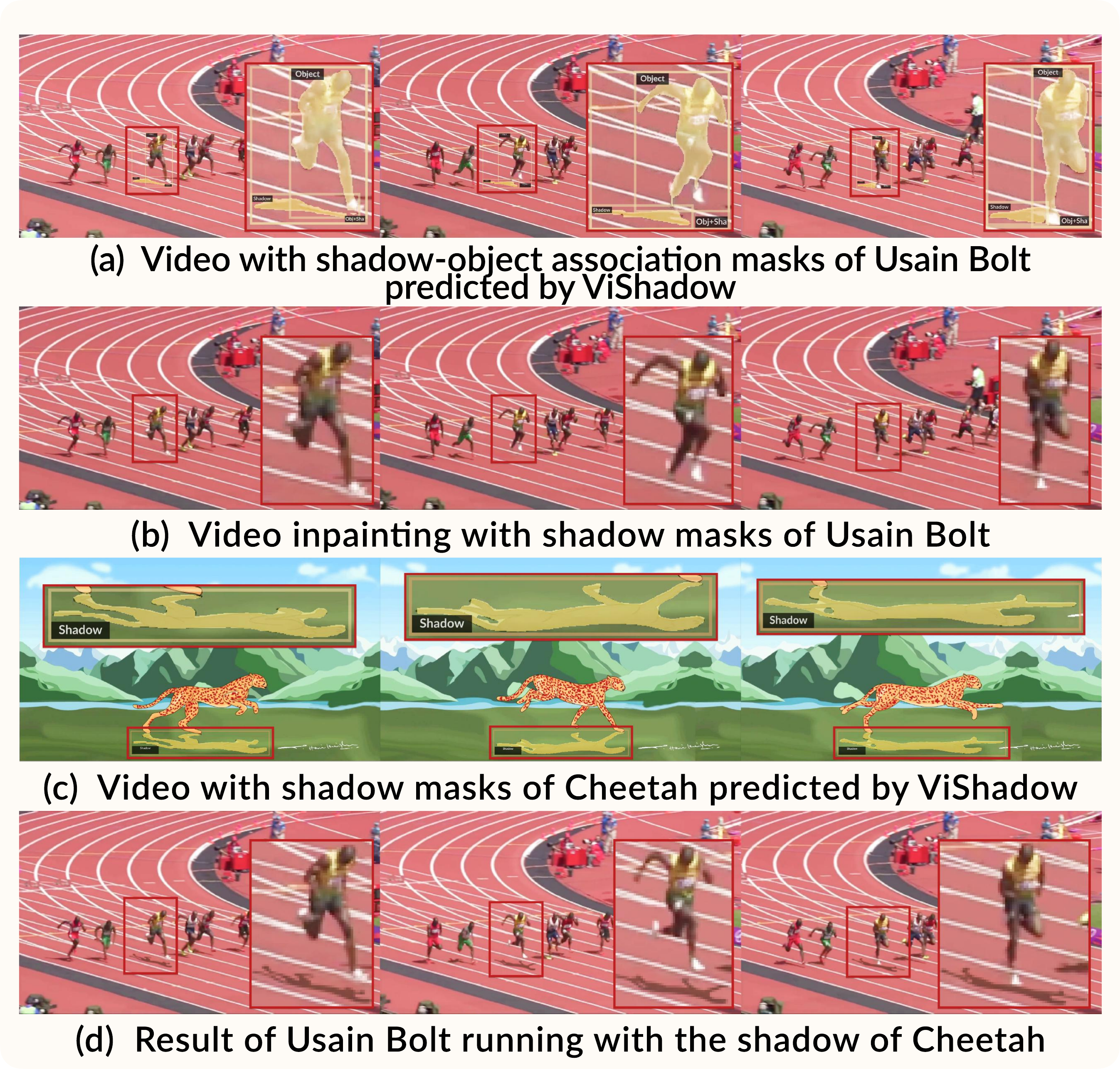}
	\caption{Usain Bolt running with the shadow of Cheetah.}
	\label{fig:mask_visd_2}

\end{figure}

Figure~\ref{fig:mask_visd_2} illustrates the application of instance-paired tracking by replacing Usain Bolt's shadow with that of a cheetah. ViShadow's predicted paired masks facilitate Usain Bolt's shadow removal through video inpainting~\cite{li2022towards}, as seen in Figure~\ref{fig:mask_visd_2} (b). Subsequently, a sequence of cheetah shadows is acquired (Figure~\ref{fig:mask_visd_2} (c)) and adjusted to Usain Bolt's object instance, resulting in the final composite shown in Figure~\ref{fig:mask_visd_2} (d).
For an immersive experience, refer to the demo video in the supplementary material.

\subsection{Text-Instructed Shadow-Object Manipulation}

The emergence of modern large-scale vision-language models, exemplified by ChatGPT-4V~\cite{openai2023gpt4} and InternGPT~\cite{liu2023interngpt}, has redefined user interaction with images and videos using natural language. Despite their extensive training with images and descriptions, these models often encounter challenges in accurately capturing shadows~\cite{jie2023sam} and shadow-object relationships. Our groundbreaking video instance shadow detection overcomes this limitation, unlocking a myriad of applications, particularly when seamlessly integrated with large vision-language models.

By integrating our method with InternGPT, we introduce an innovative platform for interactive shadow-object editing in videos, with a specific focus on the pivotal step of video instance shadow detection. This platform grants users unprecedented control over shadow-object manipulation, eliminating the necessity for task-specific model development.
Operating on the foundation of large language-vision models as central controllers, our platform autonomously dissects user requests into distinct tasks and efficiently assigns them to the most appropriate models. As illustrated in Figure~\ref{fig:text_visd}, the process unfolds seamlessly. When a user requests video inpainting based on an instance description, our controller activates ViShadow for video instance shadow detection—an integral step in discerning intricate shadow-object relationships. Subsequently, ImageBind~\cite{girdhar2023imagebind} seamlessly aligns with the identified shadow-object associations through user-provided text descriptions. Finally, the flawless execution of the shadow-object removal process by ProPainter~\cite{zhou2023propainter} underscores the effectiveness of our approach on shadow-object editing.

The paper and supplementary videos present compelling examples, offering a firsthand glimpse into the capabilities of our platform.


\begin{figure*}[tp]
	\centering
	\includegraphics[width=0.99\linewidth]{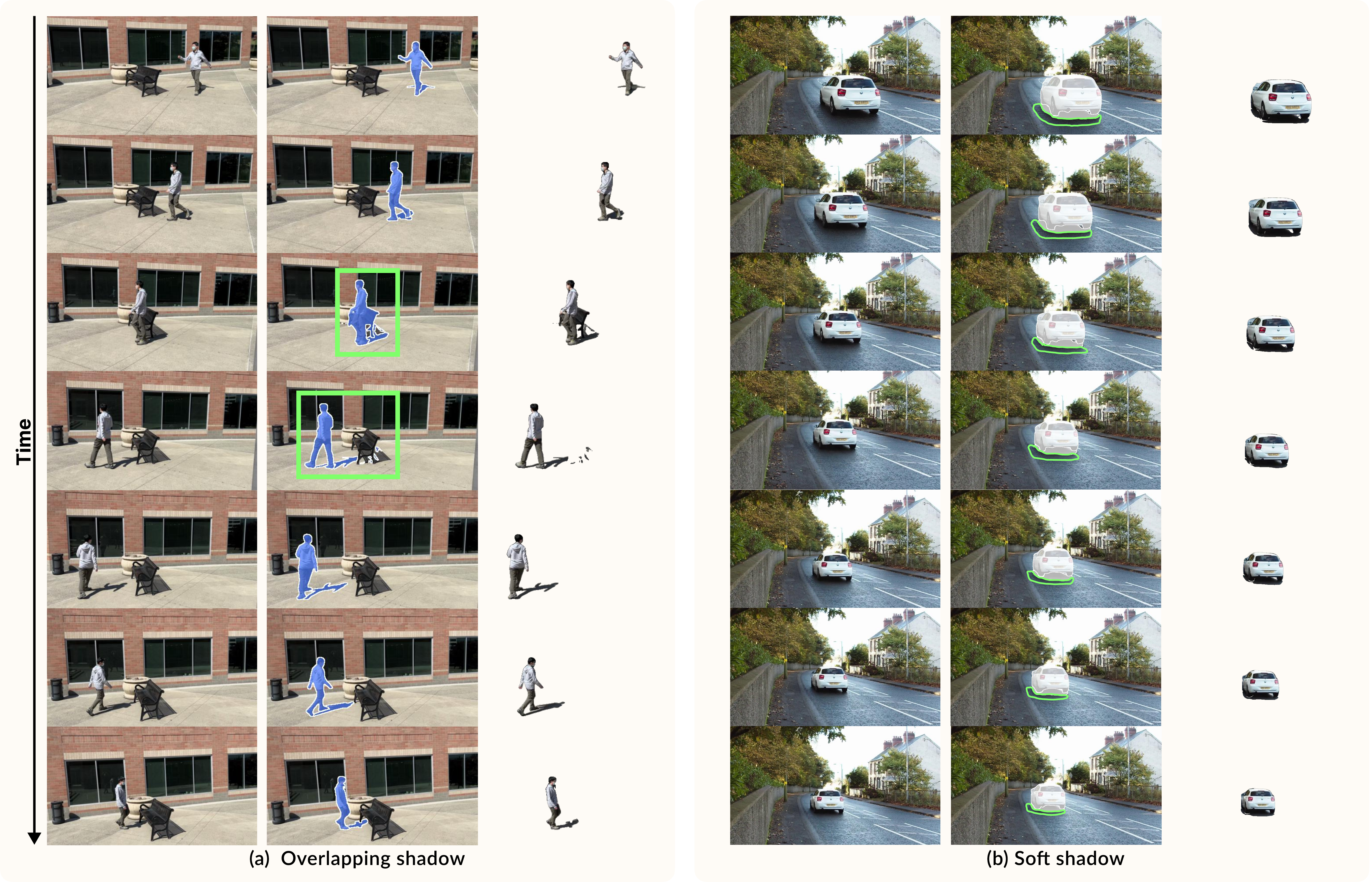}
	\caption{Left: ViShadow struggles to separate overlapping shadows, as depicted in the green boxes. Right: ViShadow encounters challenges in accurately segmenting soft shadows, as illustrated in the green boxes.}
	\label{fig:svistra_limitation}

\end{figure*}

\section{{Discussion}}
\label{sec:discussion}

\subsection{Comparison to Other Video-Related Tasks}

The tasks of video segmentation, video instance segmentation, and video shadow detection each address different aspects of video analysis but fall short in performing the specific tasks required for video instance shadow detection (VISD). Here's an exploration of why these methods are inadequate for the VISD task and the unique requirements it entails.

\textbf{Video segmentation} focuses on partitioning a video into meaningful segments based on temporal and spatial features, often labeling different regions or objects within the frames. This task can be subdivided into semantic segmentation and instance segmentation. While semantic segmentation assigns a class label to each pixel, instance segmentation goes further by distinguishing between different instances of the same class. Note that the categories of pixels or instances are predefined. 
However, video segmentation methods do not identify or associate shadows with their corresponding objects. 
They treat shadows merely as background or part of the objects, which limits their ability to handle tasks that specifically require shadow-object pairing. Also, due to the limited categories, they cannot find all objects from the videos. Moreover, standard video segmentation methods do not track the temporal continuity of shadow-object pairs across frames, which is crucial for applications like shadow editing or video inpainting where maintaining the relationship between objects and their shadows over time is essential.

\textbf{Video instance segmentation} aims to detect, segment, and track individual object instances throughout video frames. This involves maintaining the identity of each object as it moves through the video sequence, leveraging both appearance and motion cues. While video instance segmentation methods excel at tracking objects, they only focus on objects with predefined categories, and they do not specifically address shadows. 
Additionally, these methods lack mechanisms to specifically detect and associate shadows with their respective objects, thus failing to support tasks that involve shadow-object interactions.

\textbf{Video shadow detection} focuses on identifying and segmenting shadows within video frames. It aims to distinguish shadows from other elements in the scene. However, video shadow detection methods do not identify each shadow instance and typically consider shadow regions as a single category. Consequently, they cannot establish the relationship between shadows and the objects that cast them. This limitation hinders their ability to handle tasks that require understanding the context of shadow-object pairs. Moreover, these methods often do not track shadows across frames, as they lack the concept of individual shadow instances.

\textbf{Why these methods cannot perform the VISD task.}
VISD requires detecting, segmenting, associating, and tracking paired shadow-object instances in videos, which traditional methods cannot achieve. They lack the ability to establish clear shadow-object associations, maintain temporal continuity, and handle complex scenarios with occlusions and disappeared instances. 
Our framework, ViShadow, addresses these challenges with a semi-supervised approach using both labeled image data and unlabeled video data. It features a two-stage training pipeline and a retrieval mechanism for consistent tracking, enabling robust solutions for tasks like shadow removal, video inpainting, and shadow manipulation beyond the scope of traditional methods.

\subsection{Limitations}
This approach has certain limitations. First, distinguishing overlapping shadow instances from different objects is challenging because of their merged and indistinct boundaries.
Second, accurately segmenting soft shadows, characterized by gradual edges and subtle intensity transitions, poses a significant challenge. 
Refer to Figure~\ref{fig:svistra_limitation} for visual examples illustrating the challenges associated with overlapping and soft shadows.
To address these limitations in future work, inspired by \cite{chen2024learning}, we plan to use simulators, such as the game GTA-V, to render the overlapping and soft shadows, where the ground truth annotations can be generated automatically. This will expand the training and test datasets with more complex shadow interactions.

\section{Conclusion}
\label{sec:conclusions}

This paper introduces the video instance shadow detection task, aiming to detect, segment, associate, and track paired shadow-object instances in videos. We present the ViShadow framework, employing supervised learning on labeled images and self-supervised learning on unlabeled videos to learn paired tracking of shadow and object instances. A bidirectional retrieving mechanism is designed for robust paired tracking, even in cases where objects or shadows are temporarily missed. We contribute the SOBA-VID dataset for training and testing.
Experimental results demonstrate the effectiveness of our approach across various metrics. Practical applications of video instance shadow detection in video-level editing scenarios are illustrated.

\section*{Acknowledgement}
The work was supported by the National Key R\&D Program of China (Grant No. 2023YFE0202700), the Research Grants Council of the Hong Kong Special Administrative Region, China (Grant No. 14201321), and the Hong Kong Innovation and Technology Fund (Grant No. MHP/092/22).
Zhenghao Xing and Tianyu Wang are the joint first authors.

\bibliographystyle{IEEEtran}
\bibliography{IEEEtran}

\vfill

\end{document}


\title{Supplementary Material: Video Instance Shadow Detection
Under the Sun and Sky}


\onecolumn
\maketitle

\vspace*{2mm}
\large
There are four parts in this supplementary material:
\begin{description}
    \item[\textbf{Part 1} ]  \quad presents additional applications of our ViShadow.
    \item[\textbf{Part 2} ]  \quad presents more qualitative results of our ViShadow.
    \item[\textbf{Part 3} ]  \quad presents additional visual comparisons between our ViShadow and the baseline
                             \break
                             \indent 
                             \quad (SSIS+Mask2Former) [1], [2].
    \item[\textbf{Part 4} ]  \quad presents the test set and training set examples in the SOBA-VID dataset.
\end{description}

For an in-depth view of ViShadow's capabilities, please watch our supplementary video at 

https://youtu.be/uloJr5QlxR8.

\clearpage

\onecolumn

\vspace*{2mm}
\large
\noindent
{\bf Part 1: Additional Applications of Video Instance Shadow Detection}


\begin{figure*}[!htb]
	\centering
	\includegraphics[width=0.95\linewidth]{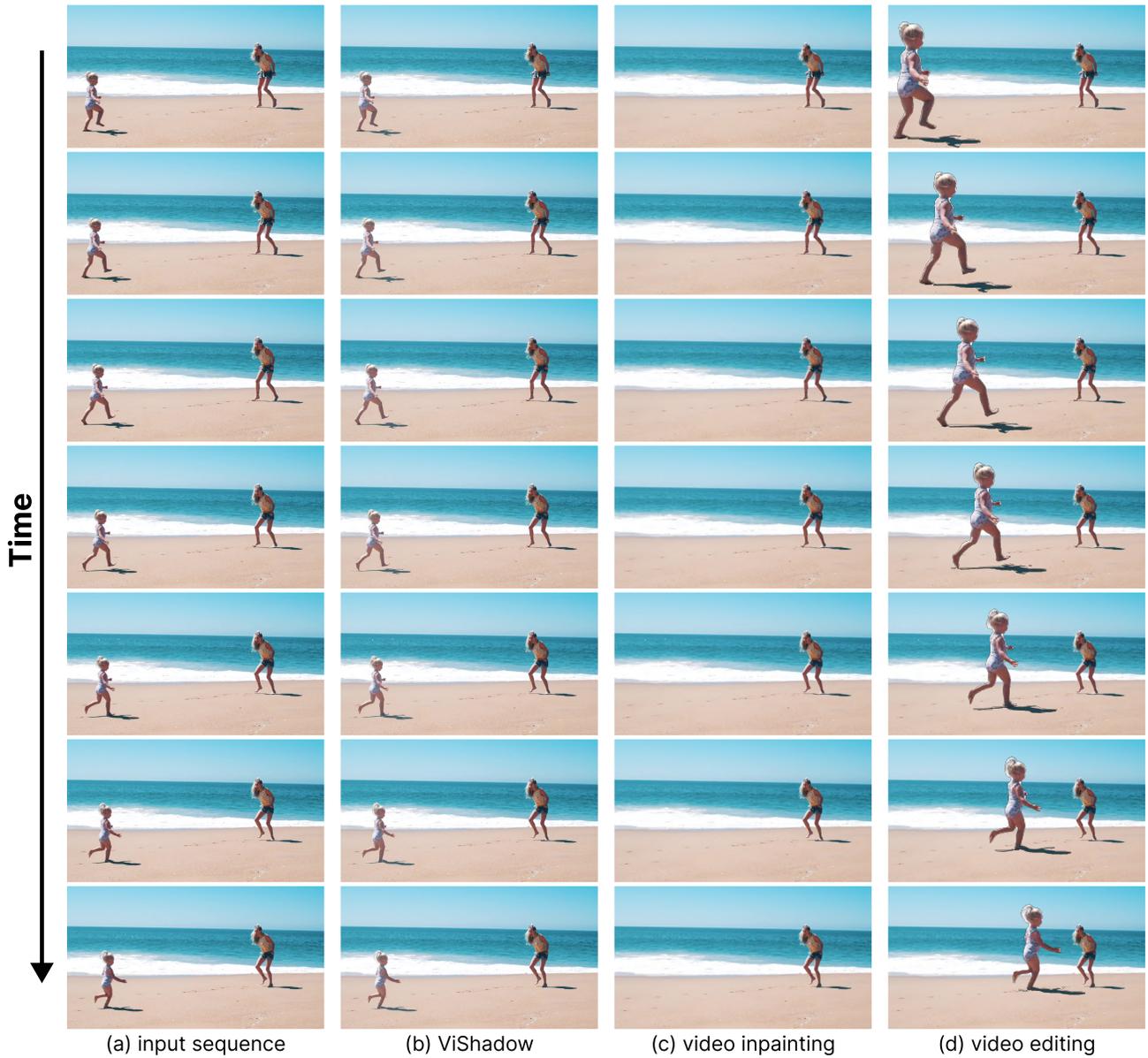}
	\caption{Example sequences demonstrate the application of video instance shadow detection in video editing. With the shadow-object associations produced by ViShadow and video inpainting algorithm [3], we can generate effects of a huge baby running to its mother.}
	\label{fig:supp_app1}
\end{figure*}

\begin{figure*}[!htb]
	\centering
	\includegraphics[width=0.95\linewidth]{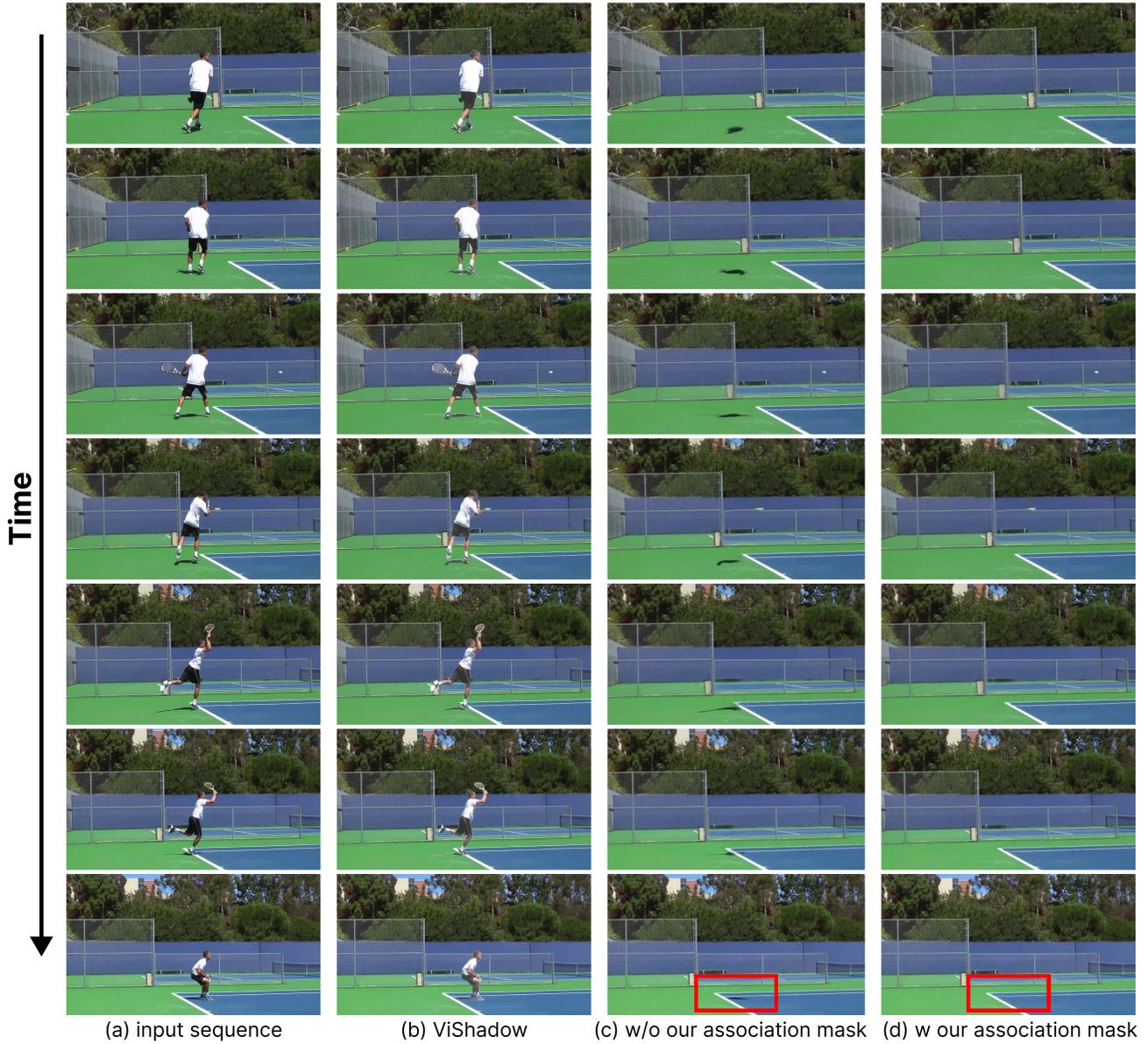}
	\caption{Example sequences demonstrate the application of video instance shadow detection in video editing. With the shadow-object associations produced by ViShadow and video inpainting algorithm [3], we can remove the tennis player and his associated shadow altogether.}
	\label{fig:supp_app2}
\end{figure*}

\clearpage

\vspace*{2mm}
\large
\noindent
{\bf Part 2: Additional Qualitative Results of ViShadow}

\vspace{5mm}

\noindent

\begin{figure*}[!htb]
	\centering
	\includegraphics[width=0.95\linewidth]{SuppFigs/more_result_1.pdf}
	\caption{Qualitative results of ViShadow.}
	\label{fig:more_result_1}
\end{figure*}

\begin{figure*}[!htb]
	\centering
	\includegraphics[width=0.95\linewidth]{SuppFigs/more_result_2.pdf}
	\caption{Qualitative results of ViShadow.}
	\label{fig:more_result_2}
\end{figure*}

\begin{figure*}[!htb]
	\centering
	\includegraphics[width=0.95\linewidth]{SuppFigs/more_result_3.pdf}
	\caption{Qualitative results of ViShadow.}
	\label{fig:more_result_3}
\end{figure*}

\begin{figure*}[!htb]
	\centering
	\includegraphics[width=0.95\linewidth]{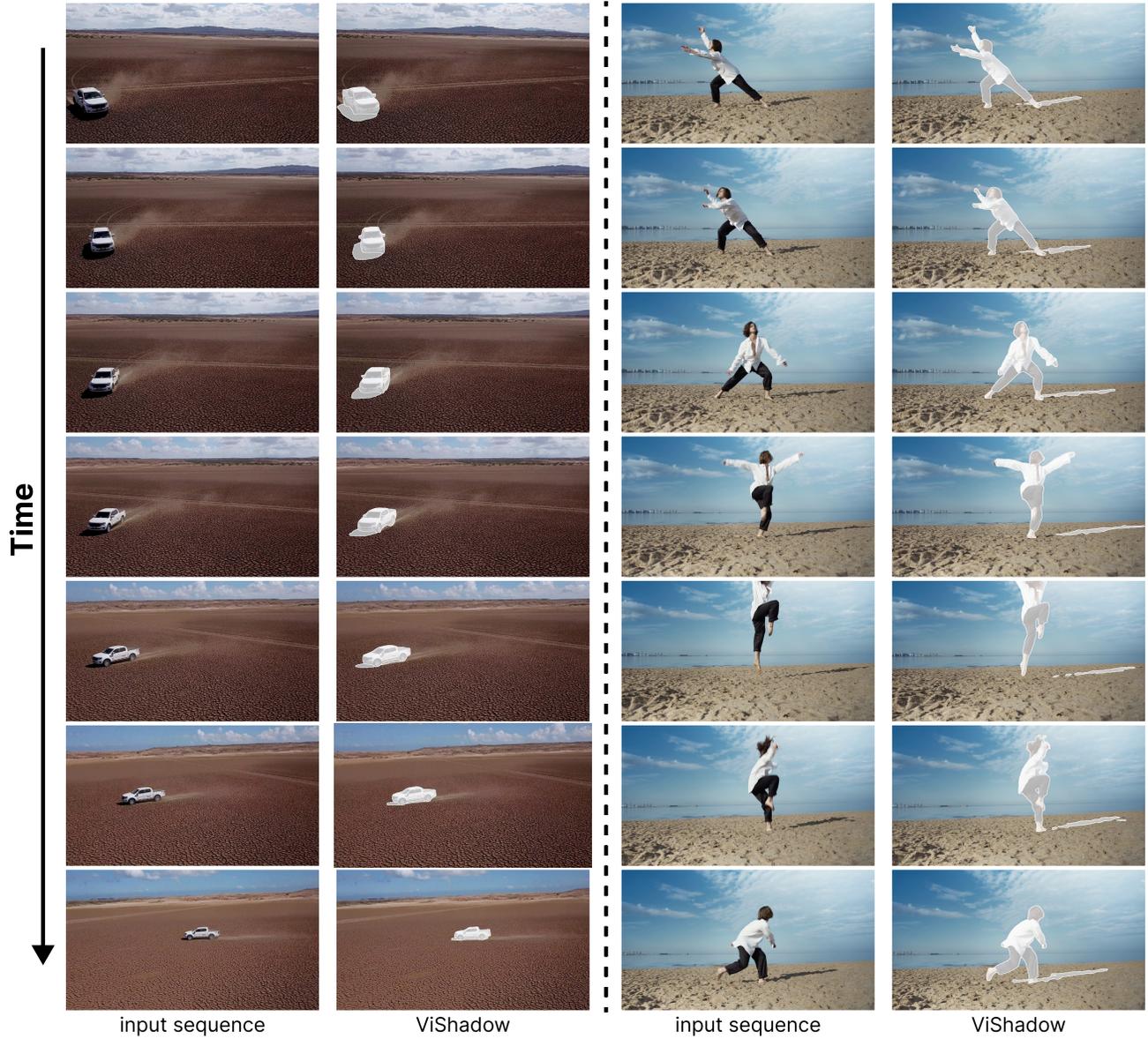}
	\caption{Qualitative results of ViShadow.}
	\label{fig:more_result_4}
\end{figure*}

\onecolumn
\clearpage

\clearpage
\vspace*{2mm}
\large
\noindent
{\bf Part 3: Visual Comparisons between Our ViShadow and the Baseline}


\begin{figure*}[!htb]
	\centering
	\includegraphics[width=0.95\linewidth]{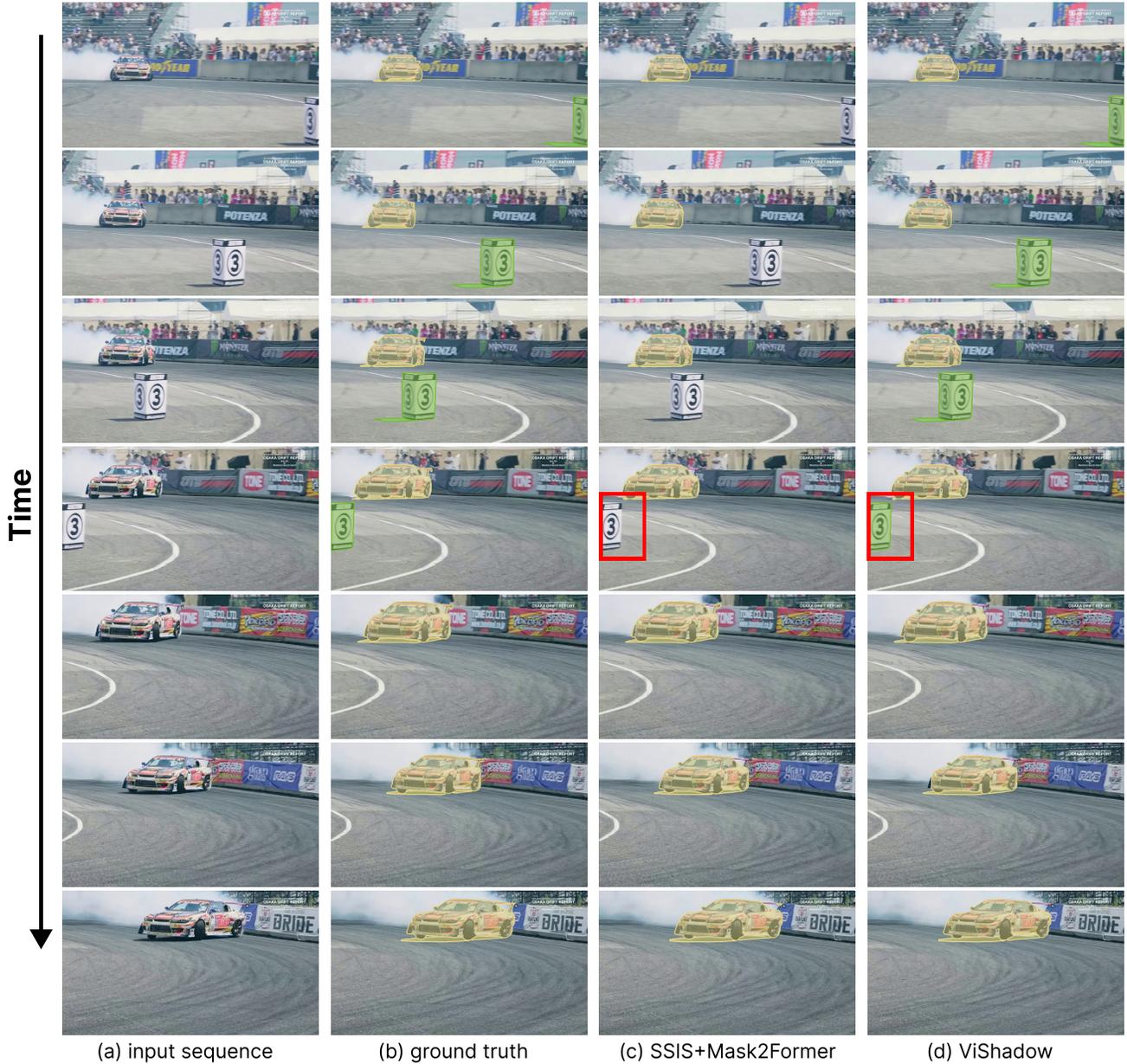}
	\caption{Visual comparison of video instance shadow detection results on sequences from the SOBA-VID test set produced by the baseline (c) and our ViShadow (d). Mask2Former [1] in the baseline cannot recognize and track the box  (marked by the red boxes above), as this category is not included in its training data.}
	\label{fig:supp_test1}
\end{figure*}

\begin{figure*}[!htb]
	\centering
	\includegraphics[width=0.95\linewidth]{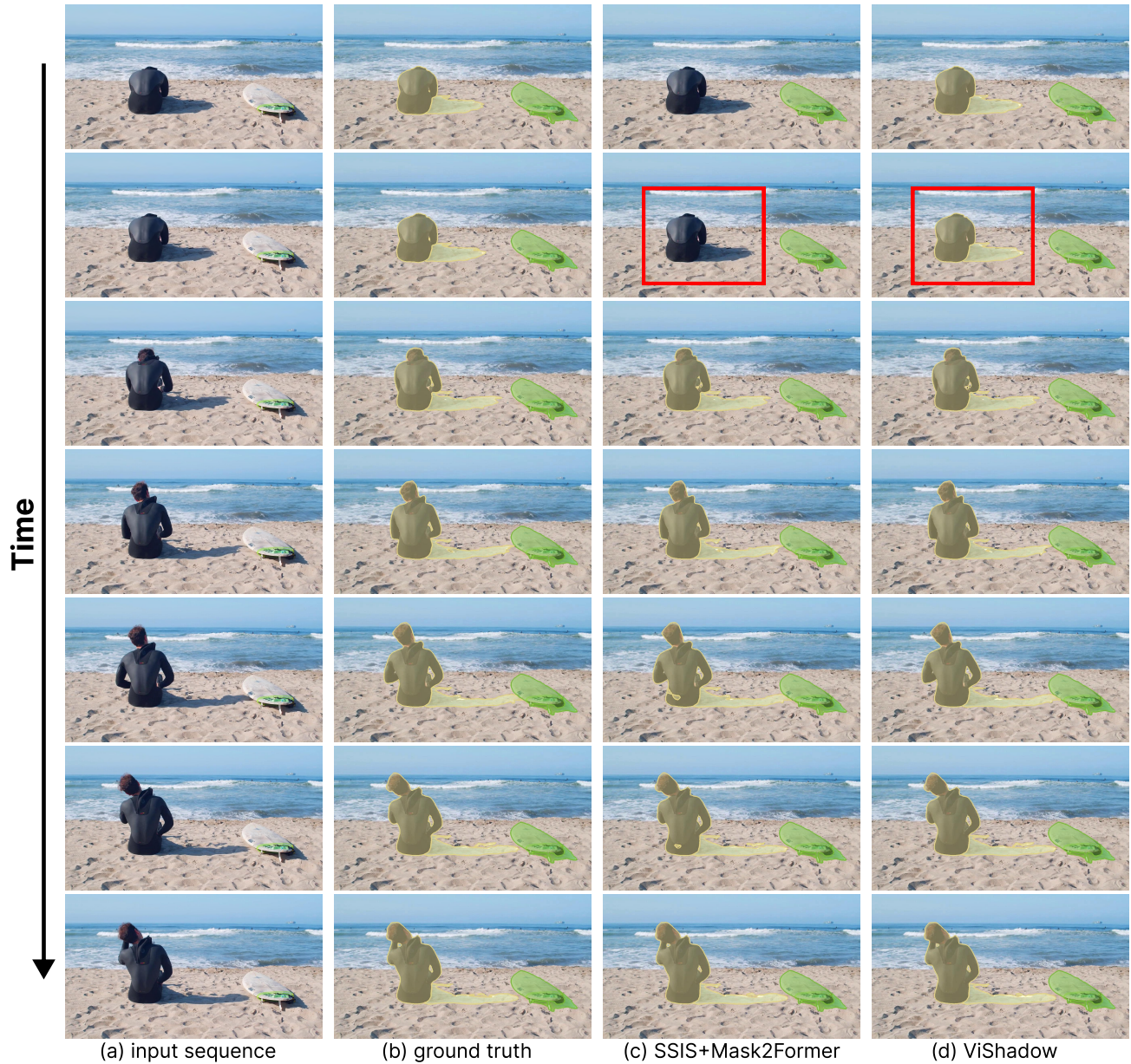}
	\caption{Visual comparison of video instance shadow detection results on sequences from the SOBA-VID test set produced by the baseline (c) and our ViShadow (d). Mask2Fomer [1] in the baseline cannot recognize and track the man in the first few frames (marked by the red boxes above).}
	\label{fig:supp_test2}
\end{figure*}

\begin{figure*}[!htb]
	\centering
	\includegraphics[width=0.95\linewidth]{SuppFigs/VIS_20004.pdf}
	\caption{Visual comparison of video instance shadow detection results on sequences from the SOBA-VID test set produced by the baseline (c) and our ViShadow (d). Mask2Former [1] in the baseline cannot recognize and track the bench (marked by the red boxes above), as this category is not included in its training data.}
	\label{fig:supp_test3}
\end{figure*}

\begin{figure*}[!htb]
	\centering
	\includegraphics[width=0.95\linewidth]{SuppFigs/VIS_90001.pdf}
	\caption{Visual comparison of video instance shadow detection results on sequences from the SOBA-VID test set produced by the baseline (c) and our ViShadow (d). Mask2Former [1] in the baseline cannot recognize and track the Charmander (marked by the red boxes above), as this category is not included in its training data.}
	\label{fig:supp_test4}
\end{figure*}

\begin{figure*}[!htb]
	\centering
	\includegraphics[width=0.95\linewidth]{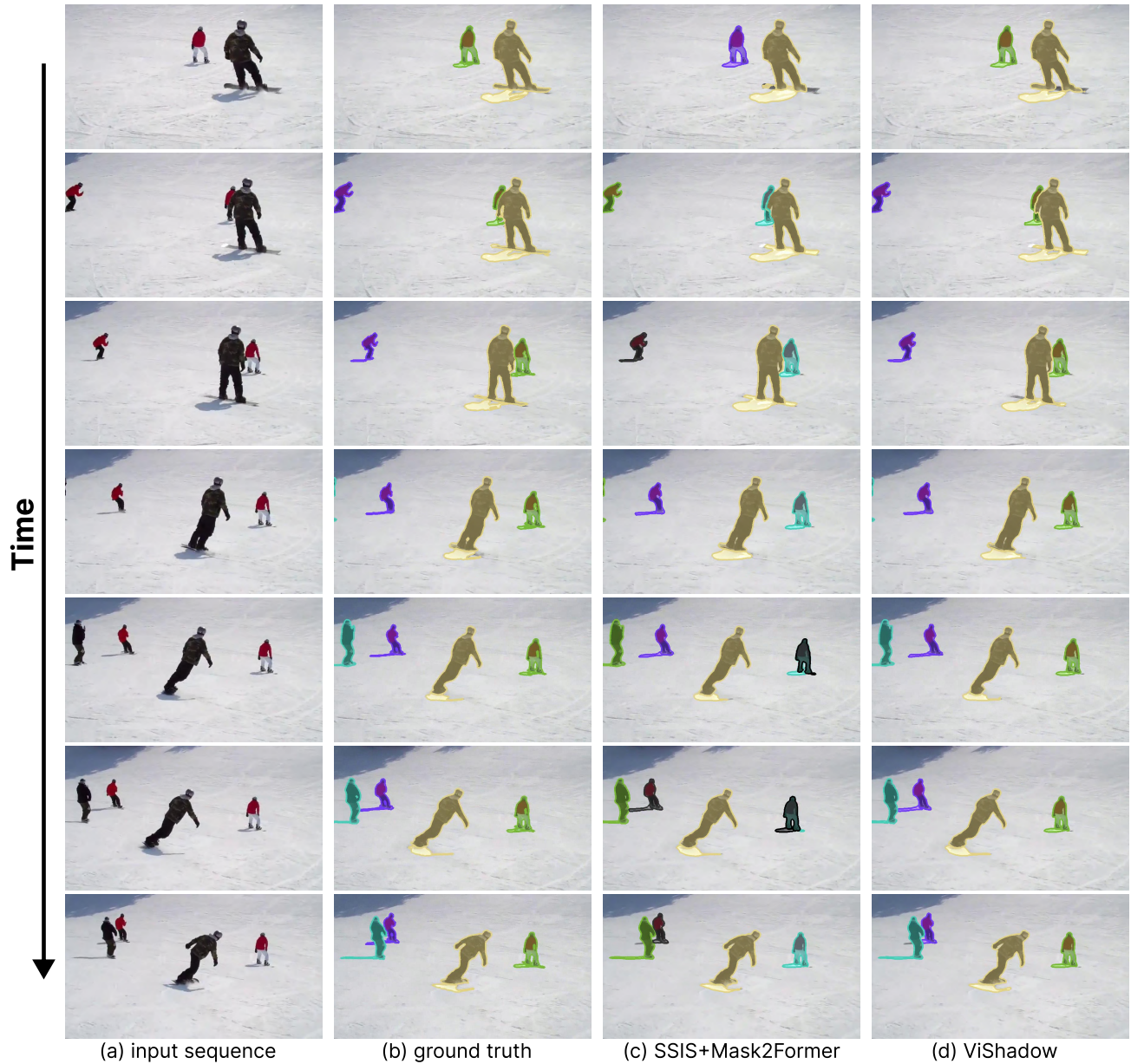}
	\caption{Visual comparison of video instance shadow detection results on sequences from the SOBA-VID test set produced by the baseline (c) and our ViShadow (d). In this case, Mask2former in the baseline cannot accurately track the instances when they sometimes occlude each other.}
	\label{fig:supp_test5}
\end{figure*}

\begin{figure*}[!htb]
	\centering
    \includegraphics[width=0.95\linewidth]{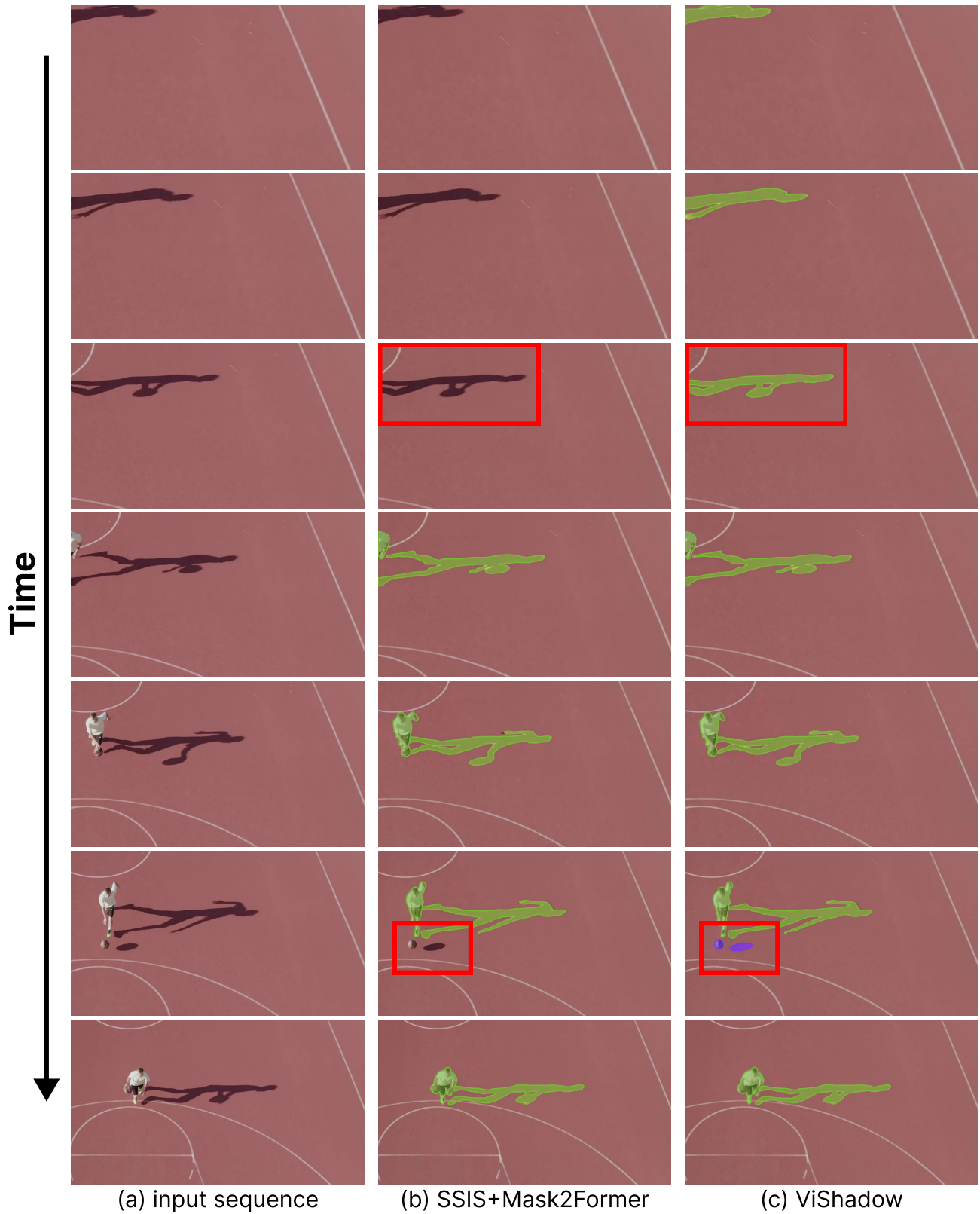}
	\caption{Visual comparison of video instance shadow detection results produced by the baseline (b) and our ViShadow (c). SSIS [2] in the baseline cannot recognize the unpaired shadow instances in the first few frames. Note that this video is from the Internet.}
	\label{fig:supp_long1}
\end{figure*}

\clearpage
\vspace*{2mm}
\large
\noindent
{\bf Part 4: Examples in the SOBA-VID Dataset}


\begin{figure*}[!htb]
	\centering
	\includegraphics[width=0.95\linewidth]{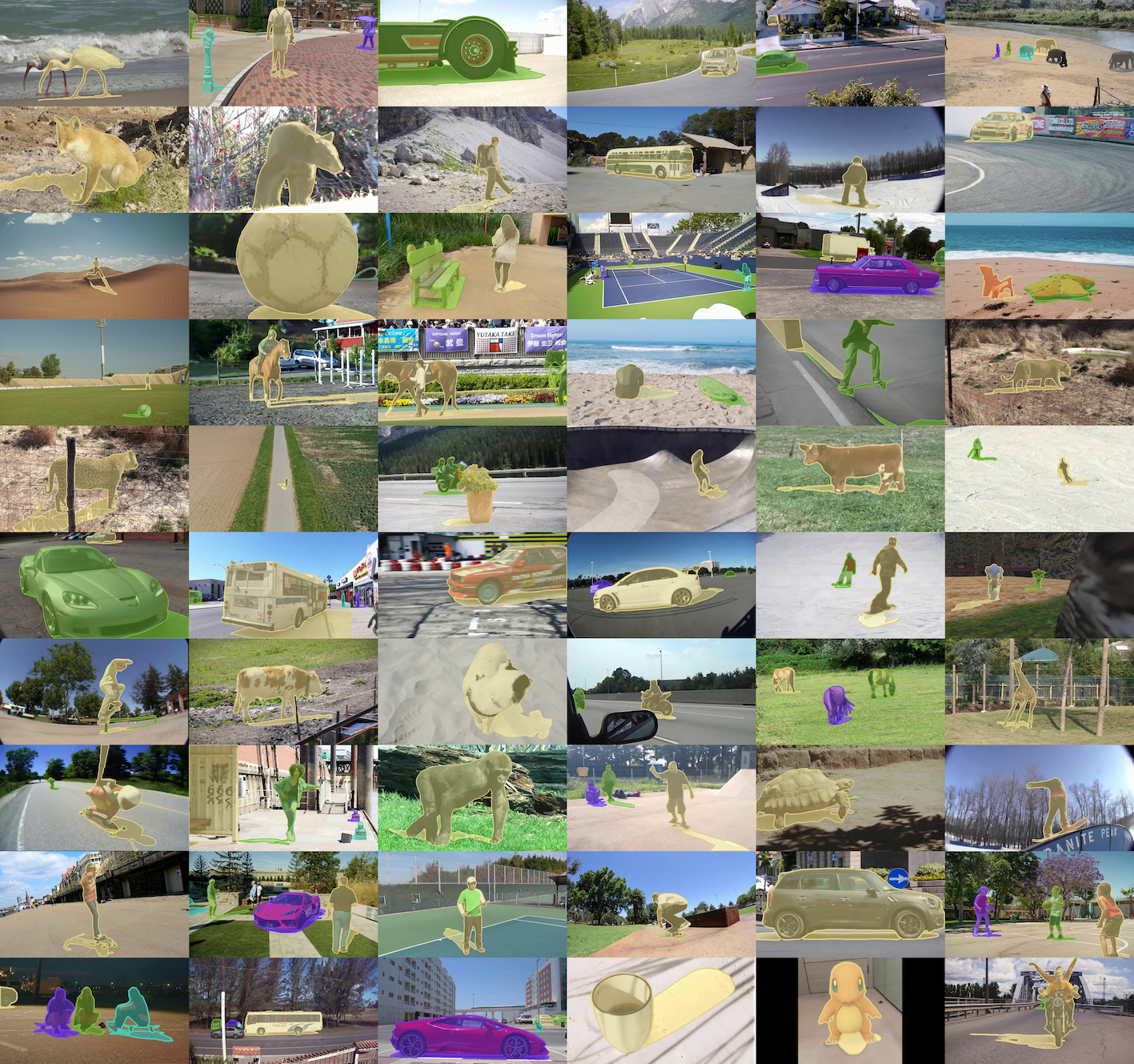}
	\caption{Examples in our SOBA-VID test set. Please zoom in for a better visualization.}
	\label{fig:supp_test_sum}
\end{figure*}

\begin{figure*}[!htb]
	\centering
	\includegraphics[width=0.95\linewidth]{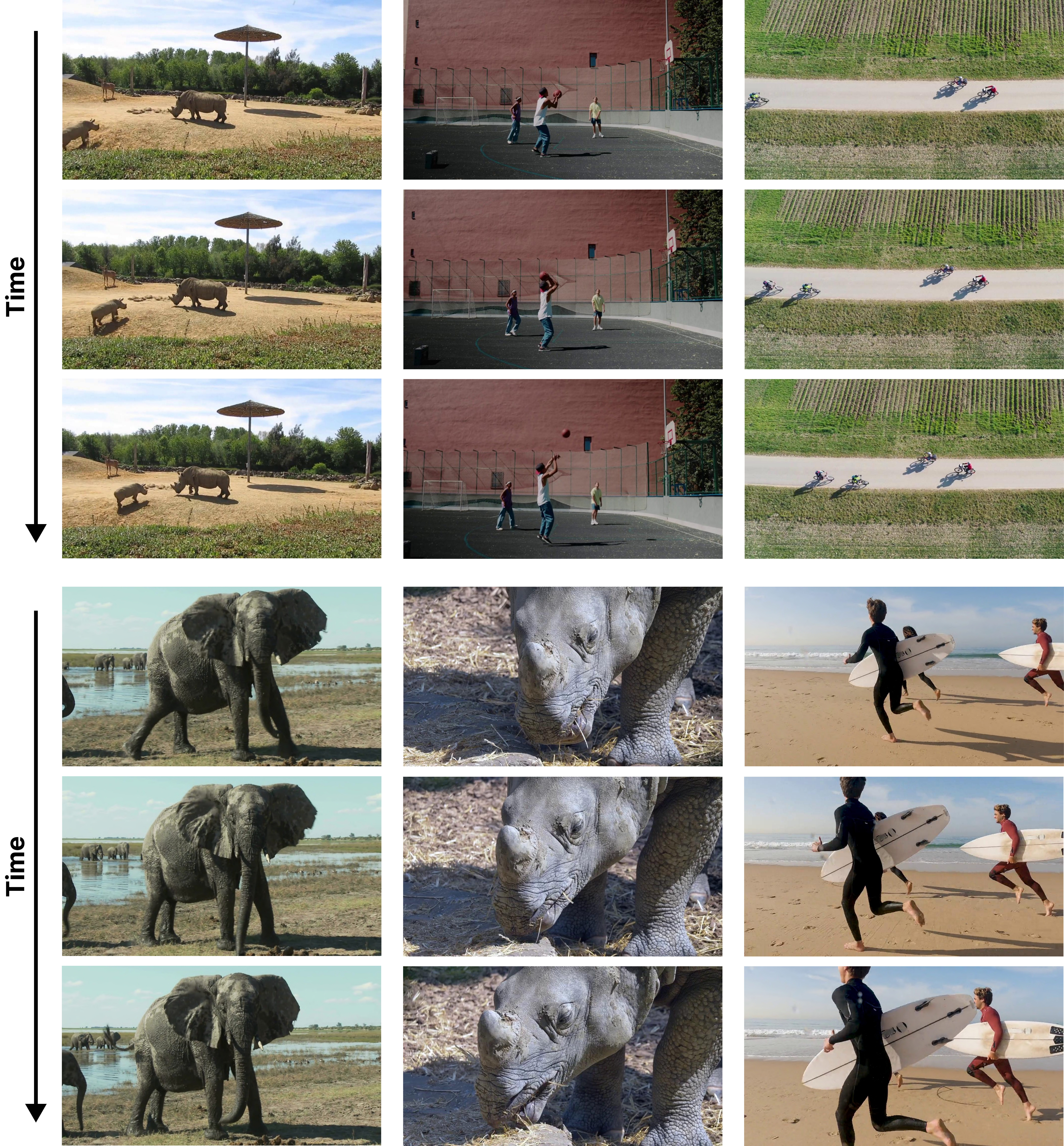}
	\caption{Video sequences in the SOBA-VID training set.}
	\label{fig:supp_train1}
\end{figure*}

\begin{figure*}[!htb]
	\centering
	\includegraphics[width=0.95\linewidth]{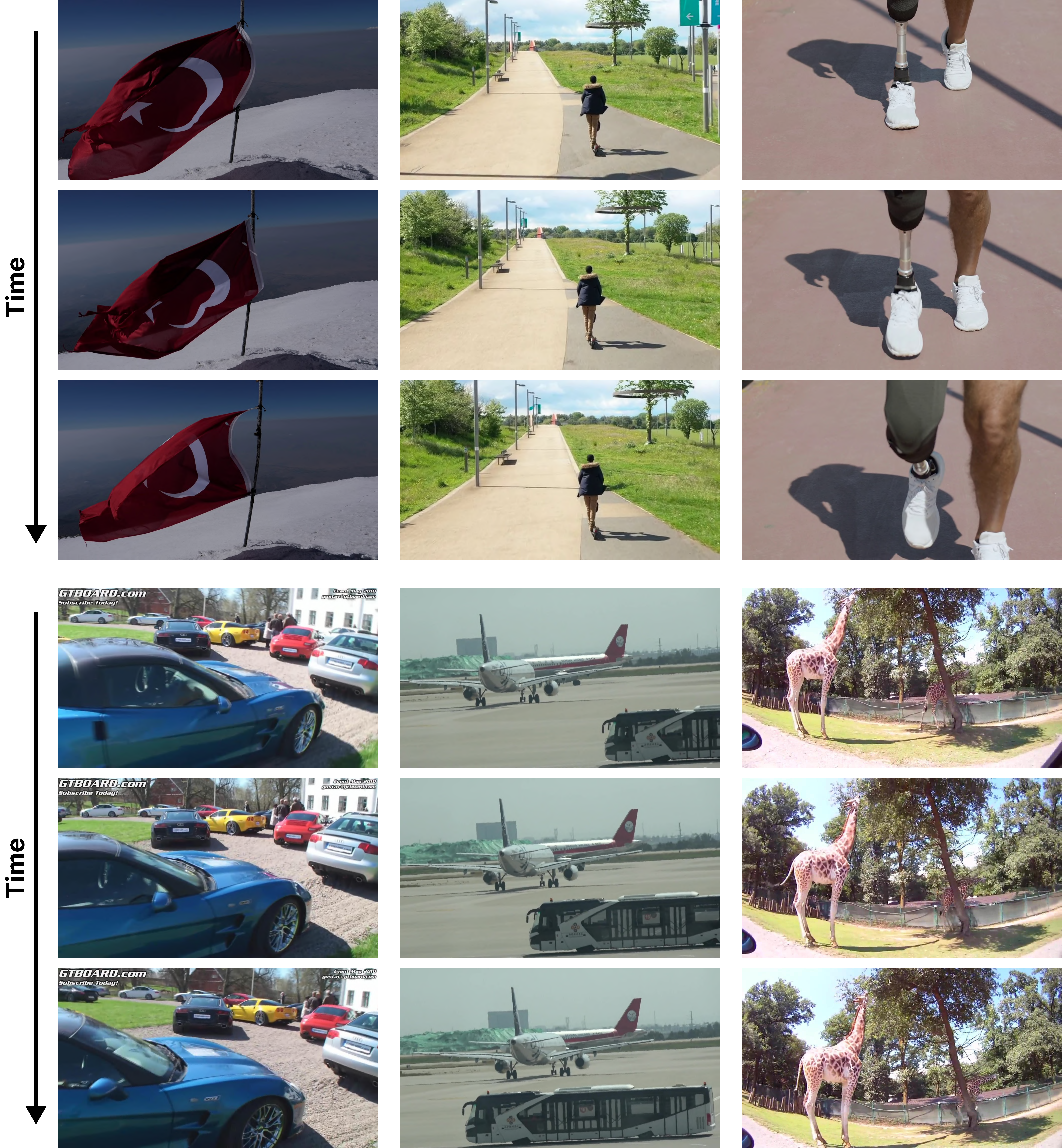}
	\caption{Video sequences in the SOBA-VID training set.}
	\label{fig:supp_train2}
\end{figure*}

\clearpage

\clearpage

{

}